\documentclass{article}

 \usepackage[preprint, nonatbib]{neurips_2024}

\usepackage[utf8]{inputenc} 
\usepackage[T1]{fontenc}    
\usepackage{hyperref}       
\usepackage{url}            
\usepackage{booktabs}       
\usepackage{amsfonts}       
\usepackage{nicefrac}       
\usepackage{microtype}      
\usepackage{xcolor}         

\usepackage{bm} 
\newtheorem{theorem}{Theorem}
\newtheorem{assumption}[theorem]{Assumption}
\newtheorem{proposition}[theorem]{Proposition}
\newtheorem{definition}{Definition} 

\usepackage{graphicx}
\usepackage{algorithm,algorithmic} 
\usepackage{amsmath} 
\usepackage{threeparttable}  
 \usepackage{subfig}
 
 \usepackage{multirow}

\title{Identification and Estimation of Long-Term \\ Treatment Effects with Monotone Missing}

\author{
  Qinwei Yang$^{1}$, Ruocheng Guo$^{2}$, Shasha Han$^{3}$, and Peng Wu$^{1}$\thanks{Corresponding author: pengwu@btbu.edu.cn.}\\ 
  $^1$Beijing Technology and Business University  \quad
  $^2$ByteDance Research   \\
 $^3$Chinese Academy of Medical Sciences
}

\begin{document}

\maketitle

\begin{abstract}
Estimating long-term treatment effects has a wide range of applications in various domains. A key feature in this context is that collecting long-term outcomes typically involves a multi-stage process and is subject to monotone missing, where individuals missing at an earlier stage remain missing at subsequent stages. Despite its prevalence, monotone missing has been rarely explored in previous studies on estimating long-term treatment effects. In this paper, we address this gap by introducing the sequential missingness assumption for identification. We propose three novel estimation methods, including inverse probability weighting, sequential regression imputation, and sequential marginal structural model (SeqMSM). Considering that the SeqMSM method may suffer from high variance due to severe data sparsity caused by monotone missing, we further propose a novel balancing-enhanced approach, BalanceNet, to improve the stability and accuracy of the estimation methods. Extensive experiments on two widely used benchmark datasets demonstrate the effectiveness of our proposed methods. 
\end{abstract}

\section{Introduction}

Estimating long-term treatment effects is of great interest in various domains. For example, economists seek to understand how early investments in education influence an individual's economic trajectory~\cite{chetty2011does}; 
 healthcare researchers investigate the long-term impacts of preventive interventions, such as vaccination programs or early screenings, on chronic diseases~\cite{chan1997human,summan2023shot}; IT companies and marketing professionals investigate the long-term user experience of mobile apps and the factors that influence retention over time~\cite{mclean2018examining}.

There are three key characteristics in estimating long-term treatment effects. 
\emph{(1) Multi-stage timeline}: collecting long-term outcomes requires an extended, multi-stage process, making them difficult to observe;  
 \emph{(2) Short-term outcomes}: before long-term outcomes become available, researchers often collect short-term outcomes as surrogates to approximate or predict long-term outcomes; 
\emph{(3) Missing data with monotone patterns}: 
both short-term and long-term outcomes are subject to missingness over time, typically following a monotone pattern in which individuals missing at an earlier stage remain missing at subsequent stages. This occurs because individuals are more likely to drop out, churn, or fail to follow-up when the preceding outcomes are not favorable~\cite{gupta2011intention, heeringa2017applied,  kallus2024role, wu2024policy}.

 However, previous studies on long-term treatment effects assume that all short-term outcomes are fully observed, overlooking the monotone missing pattern, see, e.g., \cite{kallus2024role, wu2024policy, athey2019surrogate, cheng2021long, Yang2023}.
 Monotone missing presents several special challenges. \emph{First}, it is a form of missing not at random (MNAR)~\cite{Tsiatis-2006, molenberghs2014handbook}, as the missingness of long-term outcomes depends not only on fully observed variables but also on preceding missing short-term outcomes, resulting in a difficult identifiability problem.   
 \emph{Second}, it leads to a severe data sparsity problem, as the missing ratio increases monotonically over time. As a result, the sample size of observed long-term outcomes is typically much smaller than that of missing ones, making accurate modeling and analysis highly challenging. 
 \emph{Third}, it leads to joint missing data, as when preceding outcomes are missing, the subsequent ones remain missing. How to effectively leverage the joint missing data to improve the model's performance is another key challenge.

In this article, we aim to estimate long-term treatment effects in the context of monotone missing. 
 We first introduce a novel sequential missing assumption, which is weaker than the previous missing at random (MAR) assumption. 
Then, we show the identifiability of long-term treatment effects under the sequential missing assumption and the common strongly ignorability assumption. 
In addition, we reveal that the sequential missing assumption is compatible with the monotone missing, whereas the MAR assumption is not. Subsequently, we propose two basic methods, inverse probability weighting (IPW) and sequential regression imputation (SeqRI), for estimating long-term treatment effects. 
The IPW method reweights the observed long-term outcomes based on the inverse of the product of the propensity score and the selection score. The SeqRI method imputes missing outcomes by modeling them sequentially, accounting for the dependencies between short-term and long-term outcomes.   
Unlike the traditional IPW (or regression imputation) method in causal inference~\cite{Horvitz-Thompson1952, Imbens-Rubin2015}, 
the proposed IPW (or SeqRI) method fully leverages the monotone and sequential missing mechanisms to estimate the selection score (or regression imputation) model for short/long-term outcomes.  

Moreover, a detailed analysis reveals that the IPW method relies solely on non-missing data, neglecting information from the joint missing data and thus resulting in low data utilization efficiency. 
Additionally, the SeqRI method suffers from severe model extrapolation issues. This arises because the regression imputation model trained on non-missing data must extrapolate to missing data. However, due to the MNAR and data sparsity problems, substantial differences exist between the distributions of non-missing and missing data, causing regression imputation to produce inaccurate predictions. 
Based on this analysis, we then propose the sequential marginal structural model (SeqMSM), which combines the strengths of the IPW and SeqRI methods while effectively mitigating their limitations.  
Furthermore, considering that the SeqMSM method still relies on the inverse of selection scores as weights, it may produce unstable results due to the inevitable appearance of small selection scores caused by data sparsity. To address this, we develop a novel balancing-enhanced network architecture, BalancedNet,  
which aims to learn a representation that simultaneously balances the treatment assignment and missing data. 
The main contributions are summarized below.  

~~$\bullet$  We explore and define a new monotone missing data setting for estimating multi-stage long-term treatment effects, which has a wide range of potential applications.
    
~~$\bullet$  We establish the identifiability for long-term treatment effects by introducing the sequential missing assumption, and propose three novel methods--IPW, SeqRI, SeqMSM--for estimating it. 
   
~~$\bullet$  We further develop a novel balancing-enhanced approach, BalanceNet, for improving the stability and accuracy of the proposed estimation methods.  
    
~~$\bullet$  We conduct extensive experiments to demonstrate the effectiveness of the proposed approaches.

\section{Preliminaries}
\subsection{Notation and Setup}
We consider a follow-up study where $X\in\mathcal{X}$ denotes the pre-treatment baseline covariates, $A\in\mathcal{A} = \{0,1\}$ denotes a binary treatment indicator, with $A=1$
representing treatment and $A=0$ indicating control. 
Let $S_t$ represent the short-term outcome observed at time $t$ since the start of follow-up. Denote $\bm{S}=(S_1, S_2,..., S_{t_{0}})\in\mathcal{S}\subset\mathbb{R}^{t_{0}}$ as the vector of time-varying short-term outcomes arranged in temporal order for $t_{0} \geq 2$. Let $Y\in\mathcal{Y}\subset\mathbb{R}$ be the long-term outcome of interest measured at the end of follow-up on time $T=t_{0}+1$.
In medicine, long-term outcomes usually reflect the cumulative effects of disease processes over time. For example, in an AIDS study, $S_t$ could represent biomarkers such as CD4 cell counts, viral load, or other clinical measurements collected at various times $t=1,2,..., t_{0}$, and the long-term outcome $Y$ corresponds to a definitive clinical result, such as survival status, which is observed at time $T$. Our goal is to estimate the causal effect of $A$ on $Y$. 

In the potential outcome framework~\cite{Rubin1974, Neyman1990}, we denote $Y(a)$ and $\bm{S}(a) = (S_1(a), ..., S_{t_{0}}(a))$ as the potential long-term and short-term outcomes under $A=a$ for $a = 0, 1$. The observed outcomes correspond to the potential outcomes of the actual treatment, i.e., $Y=Y(A)$, $\bm{S} = \bm{S}(A)$.

In real-world applications, as time progresses during follow-up periods, short-term and long-term outcomes generally suffer from missing due to factors such as dropout, loss to follow-up, and budget constraints~\cite{athey2019surrogate}. Let $\bm{R} = (R_1, R_2, \dots, R_{t_0}, R_T) \in \{0,1\}^T$ represent the observation indicator vector for outcomes $(\bm{S}, Y)$, where $R_t = 1$ denotes an observed outcome, and $R_t = 0$ indicates a missing one. Given the temporal order, 
we assume that individuals missing at an earlier time remain missing at later times, leading to a monotone missing pattern over time, summarized in Assumption \ref{assum:2-1}. 
\begin{assumption}[Monotone Missing]\label{assum:2-1}
For $t' > t$, if $R_t = 0$, then $R_{t'} = 0$; and if $R_{t'} = 1$, $R_t = 1$. 
\end{assumption} 
Without loss of generality, we focus on the case of $T = 3$ for simplicity, with the observed data structure presented in Table \ref{tab:1-app} of Appendix \ref{app-tmp}. The proposed methods can be easily extended to cases where $T > 3$, see Appendix \ref{Appendix:add} for a detailed discussion. 

\subsection{Problem Formulation}\label{section:2.2}
   \vspace{-8pt}

We aim to estimate the long-term average treatment effect (ATE) and conditional average treatment effect (CATE). Formally, the ATE is defined as $\tau = \mathbb{E}[Y(1)-Y(0)],$ 
which quantifies the average causal effect of the treatment on the long-term outcome in the entire population.
However, ATE only provides a rough summary of treatment effects across all units. Understanding the heterogeneity of treatment effects among different individuals is crucial for reliable decision-making in treatment evaluation selection~\cite{Chakraborty-Moodie2013, Kosorok+Laber:2019}. Therefore, we also consider the CATE, defined by  
\[
    \tau(x)= \mathbb{E}[Y(1)-Y(0)\mid X=x] := \mu_1(x)-\mu_0(x),
\] 
where $\mu_a(x) = \mathbb{E}[Y(a)\mid X=x]$ for $a=0, 1$. The CATE captures how the treatment effect varies across different subpopulations defined by features.

Due to the fundamental problem of causal inference, i.e., we can observe only one of $(Y(0), Y(1))$ for each unit. Both $\tau$ and $\tau(x)$ are not identifiable without imposing additional assumptions~\cite{pearl2009causality, Hernan-Robins2020}.  
When there are no missing values for $(\bm{S}, Y)$, the most common assumption for identifying $\tau$ and $\tau(X)$ is the following strongly ignorability assumption~\cite{pearl2009causality,rubin2005causal}.

\begin{assumption} \label{assum:2-2} 
    (a) $\{S_1(a),S_2(a),Y(a)\}\perp \!\!\! \perp A|X$; 
    (b) $0<\mathbb{P}(A=1|X=x) < 1$ for all $x\in\mathcal{X}$.
\end{assumption} 
Assumption~\ref{assum:2-2}(a), also known as the unconfoundedness assumption, holds if all confounders affecting both treatment and potential outcomes are included in the features $X$. Assumption \ref{assum:2-2}(b) asserts that for any features $X$, each treatment is assigned with a positive probability.  
Under Assumption~\ref{assum:2-2}, if $(\bm{S}, Y)$ are not subject to missing data, then $\tau(x)$ is identified as $\mathbb{E}[Y | X=x, A=1] - \mathbb{E}[Y | X=x, A=0]$, and $\tau$ is identified as $\mathbb{E} [ \mathbb{E}[Y | X=x, A=1] - \mathbb{E}[Y | X=x, A=0] ]$. However, Assumptions~\ref{assum:2-1} and \ref{assum:2-2} are insufficient to identify $\tau(x)$ and $\tau$ in the presence of missing values without further assumptions about the missing mechanisms.

\section{Challenges and Classical Method}
\subsection{Challenges} \label{sec3-1}
The sequential monotone missing pattern poses significant challenges in identifying and estimating long-term treatment effects. The challenges are threefold: 

 {\bf Missing not at random (MNAR)}.
    Both short-term and long-term outcomes are post-treatment variables, where preceding outcomes can influence both the values and the missing rates of subsequent outcomes. In other words, $R_3$ depends not only on observed variables $(X, A)$, but also on missing variables $(S_1, S_2)$, leading to the problem of MNAR~\cite{molenberghs2014handbook}. 
The MNAR problem presents a unique challenge for identifying long-term treatment effects and leads to substantial differences between the proposed methods and existing approaches for estimating these effects.

  {\bf Data sparsity}. The monotone missing mechanism implies that the sample size for observed long-term outcomes may be significantly smaller than that for missing ones. This amplifies the differences between the observed and missing data, exacerbating the missing data problem.  

 {\bf Data utilization efficiency}. 
The short-term outcomes $(S_1, S_2)$ influence both the value and the missing rate of $Y$; however, these short-term outcomes are also subject to missingness. How to effectively leverage short-term outcomes remains a key challenge in improving the precision of long-term treatment effect estimation.

\subsection{Classical Method: Missing At Random}

The classical approach to addressing missing data assumes a missing at random (MAR) mechanism, where the probability of missingness depends solely on the \emph{observed} variables~\cite{Hernan-Robins2020, zhao2024covariate}.

\begin{assumption}[Missing At Random]\label{assum:3-2}  
(a) 
$R_1\perp \!\!\! \perp S_1(a)|X,A=a; R_{2}\perp \!\!\! \perp S_{2}(a)|X,A=a;  R_{3}\perp \!\!\! \perp Y(a)|X,A=a.
$
~(b) $0<\mathbb{P}(R_t=1|X=x,A=a)<1$ for all $t$ and $a$. 
\end{assumption}

Assumption~\ref{assum:3-2}(a) is equivalent to the condition 
 expressed as $R_{1} \perp \!\!\! \perp S_{1} | X, A$, $R_{2} \perp \!\!\! \perp S_{2} | X, A$, and 
 $R_{3} \perp \!\!\! \perp Y | X, A$.  Assumption~\ref{assum:3-2}(b) implies that each outcome has a positive probability of being observed given $(X, A)$.  
Under Assumptions~\ref{assum:2-2} and \ref{assum:3-2}, the CATE $\tau(X)$ is identified as  
\begin{equation}\label{eq-1}
    \mathbb{E}[Y|X,A=1,R_3=1]-\mathbb{E}[Y|X,A=0,R_3=1].
\end{equation}
There are two main limitations of the classical method:

First, Assumption~\ref{assum:3-2} is incompatible with the monotone missing pattern given in Assumption \ref{assum:2-1}. Specifically, Assumption~\ref{assum:3-2} implies that all outcomes $(\bm{S}, Y)$ are missing at random (MAR). This assumption often does not hold in practical scenarios~\cite{chen2018social,kang2015missing}. In fact, the monotone missing pattern in Assumption~\ref{assum:2-1} suggests that the preceding outcomes may affect the missingness of subsequent outcomes. This is because units are more likely to discontinue, experience churn, or fail to participate in follow-ups when preceding outcomes are unfavorable~\cite{wu2024policy, kristman2004loss}. 

Second, low data utilization efficiency.  
The estimation based on Eq.~\eqref{eq-1} relies solely on information from $X$, disregarding the sequential structure of the missing mechanisms for $(\bm{S}, Y)$ and their potential interrelationships. In practice, the preceding short-term outcome $\bm{S}$ provides critical insights into both the missing mechanisms and the values of the long-term outcome $Y$. Moreover, Eq.~\eqref{eq-1} uses only the samples where long-term outcomes are observed, leading to losses of valuable information contained in units with missing data.  
 

Given the limitations mentioned above, we need to develop new methods to address the challenge of estimating long-term treatment effects in the monotone missing setting.

\section{Proposed IPW, SeqRI, and SeqMSM Methods}
In Section~\ref{section:4-1}, we introduce the sequential missing mechanism assumption and show the identifiability of long-term treatment effects under it. 
In Sections \ref{section:4-2}--\ref{section:4-4}, we develop three novel methods for estimating long-term treatment effects, including the inverse probability weighting (IPW), sequential regression imputation (SeqRI), and sequential marginal structural model (SeqMSM) methods.

\subsection{Identifiability}\label{section:4-1}
    

Different from  Assumption~\ref{assum:3-2} that is incompatible with the monotone missing pattern, we propose a novel sequential missing mechanism assumption below. 

\begin{assumption}[Sequential Missing Mechanism]\label{assum:4-1}
\begin{flalign}
&(a)\left\{ \hspace{-0.15cm} \begin{array}{l}
R_{3}\perp \!\!\! \perp Y(a)|X,S_1(a),S_2(a),A=a, \\
R_{2}\perp \!\!\! \perp S_{2}(a)|X,S_1(a),A=a, \\
R_{1}\perp \!\!\! \perp S_{1}(a)|X,A=a.
\end{array}\right.
\nonumber
(b) \left\{ \hspace{-0.15cm} \begin{array}{l}
r_3(X,A,S_1,S_2)=\mathbb{P}(R_3=1|X,A,S_1,S_2)>0, \\
r_2(X,A,S_1)=\mathbb{P}(R_2=1|X,A,S_1)>0, \\
r_1(X,A)=\mathbb{P}(R_1=1|X,A)>0.
\end{array}\right.&
\end{flalign}
\end{assumption}
Assumption~\ref{assum:4-1}(a) can be  equivalently expressed as $R_{3}\perp \!\!\! \perp Y | (X,A,S_1,S_2)$, $R_{2}\perp \!\!\! \perp S_{2}|(X,A,S_1)$, and $R_{1}\perp \!\!\! \perp S_{1}|(X,A)$. This implies that the outcomes $(S_2, Y)$ are MNAR, this is because the missing indicator $R_t$ ($t= 2,3$) depends not only on the observed variables $(X, A)$, but also the missing preceding outcome $S_1$. Assumption~\ref{assum:4-1}(a) also guarantees that $\mathbb{P}(Y|X, A, S_1, S_2, R_3=1)=\mathbb{P}(Y|X, A, S_1, S_2, R_3=0)$, that is, the distributions of the outcome $Y$ on the data with $R_3=1$ and the data with $R_3=0$ are comparable after accounting for $(X, A, S_1, S_2)$. Assumption~\ref{assum:4-1}(b) specifies the selection score requiring strictly positive observation probabilities for all outcomes.


Compared with Assumption \ref{assum:3-2}, Assumption \ref{assum:4-1} has several strengths: (a) it is weaker than Assumption \ref{assum:3-2}, as it allows the missing indicator to depend on preceding outcomes. As a result, 
it is more realistic and aligns with real-world scenarios, as individuals are more likely to drop out, churn, or fail in follow-up when the preceding outcomes are not desirable;     
(b) it is compatible with the monotone missing pattern, see Section \ref{section:4-3} for more details.

\begin{theorem}[Identifiability]\label{them:4-2}
Under Assumptions \ref{assum:2-1}, \ref{assum:2-2}, and \ref{assum:4-1}, $\tau$ and $\tau(x)$ are identifiable. 
\end{theorem}

Theorem \ref{them:4-2} shows the identifiability of long-term treatment effects. 
Essentially, Assumption \ref{assum:2-2} addresses the confounding bias between $A$ and $Y(a)$, while Assumptions \ref{assum:2-1} and \ref{assum:4-1} jointly tackle the MNAR problem of $Y$.  
Generally, the MNAR problem results in an unidentifiable problem unless additional assumptions are imposed, such as instrumental variables~\cite{wang2014instrumental}, shadow variables~\cite{miao2015identification}, or parametric models~\cite{miao2016identifiability}. Theorem \ref{them:4-2} extends this class of methodologies by combining the monotone missing pattern and the sequential missing mechanism.

\subsection{Inverse Probability Weighting}\label{section:4-2}

For $r_3(X, A, S_1, S_2)$ defined in Assumption \ref{assum:4-1}, one can show that (see Appendix \ref{proofs-app} for proofs) 
 \begin{equation} \label{eq2}
  \mu_a(x) = \mathbb{E} \left [ \frac{\mathbb{I}(A=a) R_3 Y}{ e_a(X) r_3(X, A, S_1, S_2) } \Big | X=x  \right],a=0,1,
\end{equation}
where the propensity score $e_a(X)=\mathbb{P}(A=a |X)$ is an identifiable quantity. Thus, if the selection score $r_3(X, A, S_1, S_2)$ is identifiable, we can construct the IPW estimator for $\tau(x)$. 
However, due to the missingness of $(S_1, S_2)$, we cannot estimate it directly by regressing $R_3$ on $(X, A, S_1, S_2)$ based on all data. 
One may estimate it directly based only on the sample with $R_{2}=1$ to eliminate missing data of $(S_1, S_2)$. If we do so, we are estimating $\mathbb{P}(R_3=1|X,A,S_1,S_2,R_{2}=1)$ instead of $r_3(X,A,S_1,S_{2})=\mathbb{P}(R_3=1|X,A,S_1,S_{2}).$ 


Fortunately, we can obtain a consistent estimator of $r_3(X,A, S_1, S_2)$ by combining the monotone missing mechanism (Assumption~\ref{assum:2-1}) and the sequential missing mechanism (Assumption  \ref{assum:4-1}). 
$$\begin{aligned}
 &  r_{3}(X,A, S_1, S_2) = \mathbb{P}\left(R_{3}=1 | X,A, S_1, S_2\right) 
= \mathbb{P}\left(R_{3}=1, R_{2}=1 | X,A, S_1, S_2\right) \\
={}& \mathbb{P}\left(R_{3}=1 | X,A, S_1, S_2, R_{2}=1\right)  \mathbb{P}\left(R_{2}=1 | X,A, S_1, S_2\right) \\
={}& \mathbb{P}\left(R_{3}=1 | X,A, S_1, S_2, R_{2}=1\right)  \mathbb{P}\left(R_{2}=1| X,A, S_1\right) \\
={}& \mathbb{P}\left(R_{3}=1 | X,A, S_1, S_2, R_{2}=1\right) \times \mathbb{P}\left(R_{2}=1 | X,A, S_1,R_1=1\right)\times \mathbb{P}(R_1=1\mid X,A),
\end{aligned}$$
where all components in the final term of the above equation are identifiable. 
Based on this, we have the following Proposition~\ref{prop:4.3}.

\begin{proposition}\label{prop:4.3} 
For the units of $R_2 = 1$,  
    $r_{3}(X,A, S_1, S_2)$ is identifiable. 
\end{proposition}

From Proposition~\ref{prop:4.3}, we can obtain consistent estimator of $r_{3}(X,A, S_1, S_2)$ by modeling the three terms $\mathbb{P}(R_1=1| X,A)$, 
$\mathbb{P}\left(R_{2}=1| X,A, S_1,R_1=1\right)$, and 
$\mathbb{P}\left(R_{3}=1 | X,A, S_1, S_2, R_{2}=1\right)$. The product of these three estimated terms is the estimate of $r_{3}(X,A, S_1, S_2)$, denoted by $\hat r_{3}(X,A, S_1, S_2)$. Similarly, denote $\hat r_{2}(X,A, S_1)$ as the estimate of $r_{2}(X,A, S_1)$, which is the product of the estimated terms $\mathbb{P}(R_1=1| X,A)$ and 
$\mathbb{P}\left(R_{2}=1| X,A, S_1,R_1=1\right)$. Denote $\hat r_{1}(X,A)$ as the estimate of $r_{1}(X,A)=\mathbb{P}(R_1=1\mid X,A)$. 
Let $\hat e_a(X)$ be the estimator of $e_a(X)$, then
 based on Eq. \eqref{eq2},  we can estimate $\mu_a(x)$ by regressing $\mathbb{I}(A=a)R_3 Y/ \hat e_a(X) \hat r_3(X, A, S_1, S_2)$ on $X$. Let $\hat \mu_a^{ipw}(x)$ represent the resulting estimator of $\mu_a(x)$, then 
  $\hat \tau_{ipw}(x) =\hat \mu_1^{ipw}(x) - \hat \mu_0^{ipw}(x)$
  is the IPW estimator of $\tau(x)$. 
In addition, $\hat \tau_{ipw} = \hat \mu_1^{ipw} - \hat \mu_0^{ipw}$, where
$\hat \mu_a^{ipw} = n^{-1}\sum_{i=1}^n \hat{\mu}^{ipw}_a(X_i)$ for $a=0,1$.


\subsection{Sequential Regression Imputation}\label{section:4-3}

We propose the sequential regression imputation (SeqRI) method for estimating long-term treatment effects. Specifically, 
under Assumptions~\ref{assum:2-2} and~\ref{assum:4-1},  $\mathbb{E}[ S_1 | X, A=a, R_1 = 0 ] = \mathbb{E}[ S_1 | X, A=a, R_1 = 1 ]$, $\mathbb{E}[ S_2 |  X, A=a, S_1, R_2 = 0 ] = \mathbb{E}[ S_2 | X, A=a, S_1, R_2 = 1 ]$, and $\mathbb{E}[ Y | X, A=a, S_1, S_2, R_3 = 0 ] = \mathbb{E}[ Y | X, A=a, S_1, S_2, R_3 = 1 ].$  
 Thus, due to the monotone missing pattern,  
 we can impute the outcomes $S_1, S_2,$ and $Y$ sequentially by modeling $m_{1a}(X) := \mathbb{E}[S_1 | X, A=a, R_1 = 1 ]$, $m_{2a}(X, S_1) := \mathbb{E}[ S_2 | X, A=a, S_1, R_2 = 1 ]$, and $m_{3a}(X, S_1, S_2) := \mathbb{E}[ Y | X, A=a, S_1, S_2, R_3 = 1 ]$, respectively. Let $\hat m_{1a}, \hat m_{2a}, \hat m_{3a}$ denote the estimates of $m_{1a},m_{2a}$ and $m_{3a}$, respectively. The corresponding estimation procedures are summarized in Algorithm~\ref{alg:1}. 
\begin{algorithm}
\caption{Sequential regression imputation algorithm}
\label{alg:1}
\begin{algorithmic}[1]
\STATE Imputing the missing value of $S_1(a)$ using $\hat{m}_{1a}(X)$ and regard  $S_1(a)$ as ``non-missing".
\STATE Imputing the missing value of $S_2(a)$ using $\hat{m}_{2a}(X,S_1)$ with $(X,\hat{m}_{1a}(X))$ as the input. Then we regard $S_2(a)$ as ``non-missing".
\STATE Predicting $Y(a)$ using $\hat{m}_{3a}(X,S_{1},S_{2})$ with $(X,\hat{m}_{1a}(X),\hat{m}_{2a}(X,S_1))$ as the input.
\end{algorithmic}
\end{algorithm}


After imputing the missing values by Algorithm~\ref{alg:1}, we can then estimate $\tau(x)$ with  
$
    \hat{\tau}_{seqri}(x)=\hat{\mu}_1^{seqri}(x)-\hat{\mu}_0^{seqri}(x),
$
where 
$\hat{\mu}_a^{seqri}(x)=\hat{m}_{3a}(x,\hat{m}_{1a}(x),\hat{m}_{2a}(x,s_1))$. In addition, for $\tau$, $\hat \tau_{seqri} = n^{-1}\sum_{i=1}^n[ \hat{\mu}_1^{seqri}(X_i)-\hat{\mu}_0^{seqri}(X_i)]$. 




\subsection{Sequential Marginal Structural Model}\label{section:4-4}
In this subsection, we analyze the limitations of the IPW and SeqRI methods and propose the sequential marginal structural model (SeqMSM) to mitigate these limitations. 

\subsubsection{Limitations of IPW and SeqRI Methods}\label{section:4.4.1}
The IPW method does not effectively address the challenges posed by the monotone missing discussed in Section \ref{sec3-1}.   
First, it has low data utilization efficiency. As shown in Eq. \eqref{eq2}, it relies solely on non-missing data $(R_3=1)$ to construct the model, ignoring the valuable information from missing individuals, leading to significant information loss.
Second, it fails to capture the dependency between the short-term outcomes $(S_1, S_2)$ and the long-term outcome $Y$. However, as shown in Assumption~\ref{assum:4-1}, $(S_1, S_2)$ are crucial predictors for $Y$.   
 Third, it is highly sensitive to small propensity scores and selection scores and suffers from a high variance problem~\cite{bang2005doubly, Kang-Schafer2007, Tan2007}. 
However, due to the data sparsity, small selection scores will inevitably appear.  



Compared with the IPW method, the SeqRI method imputes missing outcomes by modeling them sequentially, thereby enhancing data utilization efficiency and accounting for the dependencies between outcomes. However, it suffers from the problem of model extrapolation. Under the monotone missing data pattern and the associated data sparsity issue, a significant discrepancy typically exists between the distributions of non-missing and missing data, leading to severe selection bias~\cite{rosenbaum1983central,fitzmaurice2012applied}. Consequently, extrapolating the regression imputation model trained on non-missing data to the missing data results in inaccurate predictions.  

\vspace{-6pt}
\subsubsection{Sequential MSM Method}\label{section:4.4.2}
\vspace{-2pt}
To mitigate the limitations of the IPW and SeqRI methods, we propose the SeqMSM method that combines the strengths of these two methods. 
 The proposed SeqMSM method is based on the marginal structural model (MSM).  

\begin{definition}[MSM, \cite{Hernan-Robins2020}]\label{def:4-5}
The conditional means of potential outcomes $S_1(a),S_2(a)$ and $Y(a)$ are modeled by $f_1, f_2, f_3$, respectively, i.e.,  $\mathbb{E}[S_1(a)| X]=f_1(a,X),  \mathbb{E}[S_2(a)|  X,S_1(a)]=f_2(a,X,S_1(a)), \mathbb{E}[Y(a) |  X,S_1(a),S_2(a)]=f_3(a,X,S_1(a),S_2(a))$. 
\end{definition}

Definition~\ref{def:4-5} suggests that we can model the conditional means of potential outcomes using the features along with the preceding potential outcomes. 
Based on the MSM, the proposed SeqMSM method consists of three steps below.

\emph{Step 1}: Learn $f_1$ by minimizing 
    $
    \mathcal{L}_{1}(f_1)=\sum_a\mathbb{E}_n[\mathbb{I}(A=a)ww_1^{\dagger}\{S_1-f_1(a,X)\}]^2,
    $ 
where $\mathbb{E}_n$ represents the empirical mean operator, defined as  
$\mathbb{E}_n[U] =n^{-1}\sum_{i=1}^n U_i$ with $U$ being a generic variable. Denote $\hat f_1$ as the learned $f_1$. 

\emph{Step 2}: Learning $f_2$ by minimizing 
    $
    \mathcal{L}_{2}(f_2)=\sum_a\mathbb{E}_n[\mathbb{I}(A=a)ww_2^{\dagger}\{S_2-f_2(a,X, \hat f_1)\}]^2.
    $
Denote $\hat f_2$ as the learned $f_2$. 

\emph{Step 3}: Learn $f_3$ by minimizing 
    $
      \mathcal{L}_{3}(f_3)=\sum_a\mathbb{E}_n[\mathbb{I}(A=a)ww_3^{\dagger}\{Y-f_3(a,X,\hat{f}_1,\hat{f}_2)\}]^2,
    $
where $w =1/\hat e_a(X)$, and $w_t^{\dagger}=R_t/\hat{r}_t, t\in\{1,2,3\}$. 
Denote $\hat f_3$ as the learned $f_3$.

The SeqMSM method effectively mitigates the problem of model extrapolation by weighting the non-missing data by inverse probability of propensity score and selection score. Concretely, both the propensity score and selection score are balancing score~\cite{imai2014covariate}. By weighting the non-missing data with the inverse of them, we can create an unbiased pseudo-population that mimics the whole population. As a result, training the prediction model in the pseudo-population acts as training in the target population, thereby enhancing the model's extrapolation ability.
In addition, the SeqMSM method effectively reduces information loss and accounts for the
dependencies between outcomes. Similar to the SeqRI method, it uses a sequential regression imputation model to estimate long-term effects,  utilizing information from missing data and capturing the dependencies between outcomes.

\section{Proposed Balancing-Enhanced Network}\label{section:4.5} 

Although the SeqMSM method has several desirable merits, its reweighting mechanism relies on the inverse of both propensity and selection scores, making it susceptible to the risk of high variance.  
Considering that the IPW method aims to construct balanced feature distributions among the units with different treatments $A$ and missing patterns $R$ using inverse propensity and selection scores~\cite{rosenbaum1983central,imai2014covariate, li2023propensity}, we propose a novel balancing-enhanced neural network based approach, BalanceNet, to improve the stability of the SeqMSM method for long-term treatment effect estimation. 

\begin{figure}[!h]
    \centering
    \includegraphics[width=0.8\linewidth]{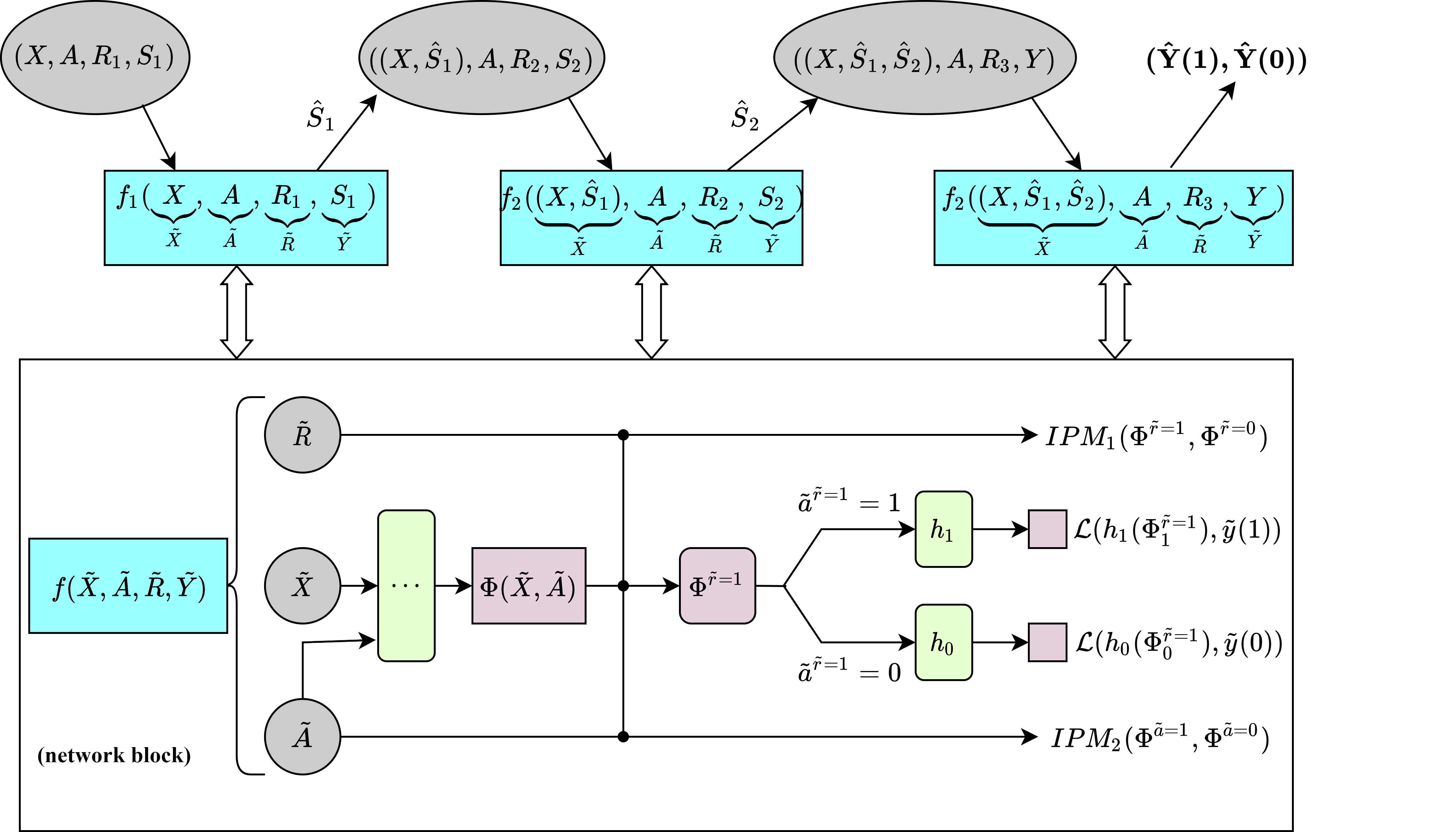}
    \caption{Architecture of BalanceNet.}
    \label{fig:1}
\end{figure}

As shown in the top panel of Figure~\ref{fig:1}, BalanceNet has a sequential structure, with each time step employing the same network block shown in the bottom panel. %
The network block is composed of two submodules. 
 First, the Encoder maps $(\tilde{X},\tilde{A})$ to the representation $\Phi(\tilde{X},\tilde{A})$. 
Second, the Outcome Predictor takes the representation $\Phi^{\tilde{r}=1}$ of the units with observed outcomes as its input and predicts outcomes. The Outcome Predictor consists of two
separate \textit{prediction heads} $h_0=h(\Phi^{\tilde{r}=1},0)$ and $h_1=h(\Phi^{\tilde{r}=1},1)$, the former (latter) estimates the outcome under control (treatment).

For balancing the representations, we employ the Integral Probability Metric (IPM)~\cite{sriperumbudur2012empirical} to measure the distance between two imbalanced distributions. Minimizing IPM would penalize imbalanced distributions of learned representations $\Phi(\tilde{X},\tilde{A})$.
In particular, we adopt the squared linear MMD distances~\cite{smola2006maximum,borgwardt2012kernel}, a type of IPM widely adopted for its computational efficiency as it can be computed in closed form with samples from unknown distributions. 
Finally, the BalanceNet consists of two penalty terms for balancing in its loss function, one is for balancing the discrepancy of the missing events and observed events by the regularization term $\textup{IPM}_1(\Phi^{\tilde{r}=1},\Phi^{\tilde{r}=0})$, and the other is balancing the discrepancy of the \textit{treated} group and the \textit{control} group by the term $\textup{IPM}_2(\Phi^{\tilde{a}=1},\Phi^{\tilde{a}=0})$. To this end, we
seek a representation $\Phi$ and hypothesis $h$ that minimizes a loss that performs a trade-off between predictive accuracy for the observed outcomes and imbalance in the
representation space, by the following objective for each time step. 
\begin{equation}\label{eq:4}
\begin{aligned}
    & \min_{h,\Phi} \sum_{i=1}^n \Big [ \mathcal{L}(h(\Phi^{\tilde{r}=1}(\tilde{X}_i,\tilde{A}_i),\tilde{Y}_i)  + \lambda_1\cdot \textup{IPM}_1(\Phi^{\tilde{r}=1}(\tilde{X}_i,\tilde{A}_i),\Phi^{\tilde{r}=0}(\tilde{X}_i,\tilde{A}_i)) \\
    {}&+ \lambda_2\cdot \textup{IPM}_2(\Phi^{\tilde{a}=1}(\tilde{X}_i),\Phi^{\tilde{a}=0}(\tilde{X}_i)) \Big ],
\end{aligned}
\end{equation}
where $\lambda_1, \lambda_2 \ge 0$ are imbalance penalties, that control the contributions of the balancing terms. We train BalanceNet by minimizing the loss function in  Eq.~\eqref{eq:4} by Adam.


\section{Experiments}
To demonstrate the proposed methods, we conduct experiments on two widely used datasets, IHDP~\cite{hill2011bayesian} and JOBS~\cite{lalonde1986evaluating}.
Detailed descriptions of these datasets are provided in Appendix \ref{Appendix:c}. 

\subsection{Experimental Setup}

{\bf Simulating Outcomes}. None of the IHDP and JOBS datasets collect short-term and long-term outcomes, we simulate short-term and long-term outcomes for both datasets to determine the ground truth of the causal effect at each timestep. The data-generating mechanisms follow previous studies~\cite{wu2024policy,cheng2021long} and are detailed in  Appendix \ref{Appendix:c} due to space constraints. 
In the data-generating mechanisms, $C_1$ and $C_2$ 
  are key constants that control the influence of historical outcomes on current outcomes.  

{\bf Missing Mechanism}. Following the missing mechanism in Assumption~\ref {assum:2-1} and Assumption~\ref{assum:4-1}. Let $\gamma$ denote the missing ratio. We simulate the missing mechanism as shown in Appendix~\ref{Appendix:c}.

{\bf Implementation Details}. The details of the model implementation are also provided in Appendix~\ref{Appendix:c}.

{\bf Baselines}. 
We compare three baseline methods in the above datasets. (1) \textbf{Naive-OR}, an outcome regression model with separate regressors for each treatment, each regressor uses features as predictors to estimate the outcomes. (2) \textbf{Naive-IPW}~\cite{caron2022estimating}, the Naive-IPW utilizes inverse propensity scores to balance feature distributions of the treated and control groups through reweighting.
(3) \textbf{CFRNet}~\cite{shalit2017estimating}, a deep-learning-based method for causal effect estimation. It predicts individual treatment effects by learning
balanced feature representations between the treated group and the control group.

{\bf Evaluation Metrics}. We report two widely used evaluation metrics for long-term causal effects estimation, including the absolute error in estimated average treatment effect  $\epsilon_{ATE} = \lvert\frac{1}{n}\sum_{j=1}^{n}(\hat{y}_j(1)-\hat{y}_j(0))-\frac{1}{n}\sum_{j=1}^{n}(y_j(1)-y_j(0)) \rvert$, and the square root of the mean square errors (RMSE) of the estimated condition average treatment effect $\epsilon_{CATE}=\sqrt{\frac{1}{n}\sum_{j=1}^{n}\{\hat{\tau}(x_j)-\tau(x_j)\}^2}$.

\begin{table*}[!t]
\centering
\caption{Performance comparison on IHDP and JOBS with $\lambda_1=\lambda_2=1,C_1=5,C_2=2,T=3$, the  variance information in Appendix~\ref{Appendix:d}.}
\resizebox{1 \linewidth}{!}{
\begin{tabular}{c|cc |cc|cc|cc|cc|cc|cc}
\toprule
\multicolumn{1}{c|}{JOBS} & \multicolumn{2}{c|}{$\gamma=0.05$}  & \multicolumn{2}{c|}{$\gamma=0.1$}   & \multicolumn{2}{c|}{$\gamma=0.15$}   
& \multicolumn{2}{c|}{$\gamma=0.2$}        & \multicolumn{2}{c|}{$\gamma=0.3$}    
& \multicolumn{2}{c|}{$\gamma=0.4$}
& \multicolumn{2}{c}{$\gamma=0.5$}
\\ \midrule
 Method & $\epsilon_{CATE}$  & $\epsilon_{ATE}$ & $\epsilon_{CATE}$  & $\epsilon_{ATE}$ & $\epsilon_{CATE}$  & $\epsilon_{ATE}$ & $\epsilon_{CATE}$ & $\epsilon_{ATE}$   &$\epsilon_{CATE}$ & $\epsilon_{ATE}$  & $\epsilon_{CATE}$  & $\epsilon_{ATE}$  & $\epsilon_{CATE}$    & $\epsilon_{ATE}$    
 \\ \midrule 

Naive-OR &2.978  & 1.064  & 4.760  & 1.433  & 4.646  & 1.748  & 4.660  & 1.343  & 4.135  & 1.566  & 3.941  & 1.486  & 3.510  & 1.620   \\
Naive-IPW &6.020  & 2.402  & 5.881  & 2.384  & 5.060  & 2.033  & 6.033  & 2.092  & 4.943  & 1.776  & 5.330  & 1.714  & 4.069  & 1.246   \\
CFRNet &1.659  & 0.727  & 1.754  & 0.895  & 1.791  & 0.990  & 1.862  & 1.122  & 2.067  & 1.325  & 2.193  & 1.370  & 2.151  & 1.429   \\
\midrule
Proposed-IPW &6.212  & 2.647  & 5.176  & 1.863  & 5.406  & 2.106  & 6.416  & 2.440  & 5.747  & 1.884  & 3.786  & 0.949  & 4.342  & 1.616   \\
SeqRI&3.186  & 0.913  & 4.823  & 1.284  & 4.883  & 1.317  & 6.957  & 2.590  & 4.696  & 1.391  & 4.387  & 1.625  & 3.995  & 1.332   \\
SeqMSM&1.461  & 0.611  & 1.537  & 0.736  & 1.613  & 0.908  & 1.712  & 1.059  & 1.755  & 1.099  & 1.772  & 1.131  & 1.797  & 1.148   \\
BalanceNet &\textbf{1.284*}  & \textbf{0.222*}  & \textbf{1.285*}  & \textbf{0.207*}  & \textbf{1.283*}  & \textbf{0.199*}  & \textbf{1.295*}  & \textbf{0.235*}  & \textbf{1.285*}  & \textbf{0.222*}  & \textbf{1.302*}  & \textbf{0.250*}  & \textbf{1.315*}  & \textbf{0.265*}   \\
  \midrule
 
  \midrule
  \multicolumn{1}{c|}{IHDP}   & \multicolumn{2}{c|}{$\gamma=0.05$}  & \multicolumn{2}{c|}{$\gamma=0.1$}   & \multicolumn{2}{c|}{$\gamma=0.15$}   
& \multicolumn{2}{c|}{$\gamma=0.2$}        & \multicolumn{2}{c|}{$\gamma=0.3$}    
& \multicolumn{2}{c|}{$\gamma=0.4$}
& \multicolumn{2}{c}{$\gamma=0.5$}
\\ \midrule
 Method &$\epsilon_{CATE}$  & $\epsilon_{ATE}$  &$\epsilon_{CATE}$  & $\epsilon_{ATE}$  &$\epsilon_{CATE}$  & $\epsilon_{ATE}$ & $\epsilon_{CATE}$ & $\epsilon_{ATE}$   &$\epsilon_{CATE}$ & $\epsilon_{ATE}$  & $\epsilon_{CATE}$  & $\epsilon_{ATE}$  & $\epsilon_{CATE}$    & $\epsilon_{ATE}$   
 \\ \midrule 

Naive-OR &8.614  & 0.914  & 8.823  & 1.565  & 9.675  & 2.168  & 10.656  & 2.759  & 13.628  & 5.464  & 14.270  & 5.144  & 14.052  & 4.991   \\
Naive-IPW &7.758  & 2.419  & 10.146  & 3.285  & 8.126  & 1.828  & 9.938  & 3.389  & 12.499  & 4.599  & 12.572  & 5.689  & 12.998  & 5.531   \\
CFRNet &10.738  & 4.230  & 10.128  & 3.609  & 10.312  & 2.878  & 8.879  & 2.279  & 9.231  & 1.971  & 9.716  & 3.045  & 10.592  & 3.411   \\
\midrule
Proposed-IPW &7.793  & 1.797  & 7.986  & 1.395  & 10.055  & 3.823  & 10.855  & 3.606  & 12.957  & 5.332  & 13.729  & 6.695  & 12.997  & 5.613   \\
SeqRI&12.983  & 3.682  & 13.435  & 4.099  & 14.328  & 2.833  & 14.540  & 3.618  & 14.963  & 4.267  & 15.488  & 4.262  & 13.592  & 2.832   \\
SeqMSM&5.770  & \textbf{0.901*}  & 5.888  & \textbf{0.985*}  & 5.668  & \textbf{1.063*}  & 6.051  & \textbf{1.459*}  & 6.241  & 2.148  & 6.084  & \textbf{1.734*}  & 6.876  & 2.478   \\
BalanceNet &\textbf{5.340*}  & 1.992  & \textbf{5.347*}  & 2.008  & \textbf{5.352*}  & 1.998  & \textbf{5.394*}  & 2.025  & \textbf{5.429*}  & \textbf{2.088*}  & \textbf{5.426*}  & 2.104  & \textbf{5.458*}  & \textbf{2.123*}   \\   
  \bottomrule
  \end{tabular}}
  \begin{tablenotes} 
    \footnotesize
    \item  Note: * statistically significant results (p-value $\leq 0.05$) using the paired t-test compared with the baseline. 
    \end{tablenotes}
\label{tab:2}%
\end{table*}%



\begin{table}
  \begin{minipage}[p]{0.65\textwidth}
    \centering
    \resizebox{\linewidth}{!}{
    \subfloat[on JOBS]{
    \begin{minipage}[t]{0.4\linewidth}
    \centering
    \includegraphics[width=1.0\textwidth]{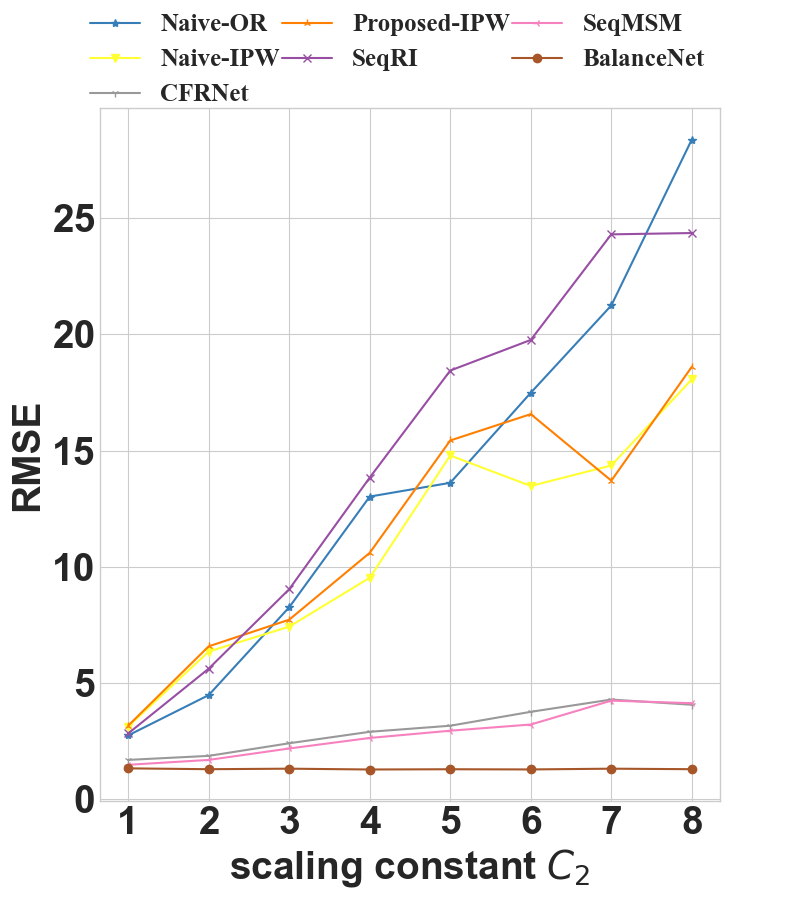}
    \end{minipage}%
    }%
    \subfloat[on IHDP]{
    \begin{minipage}[t]{0.4\linewidth}
    \centering
    \includegraphics[width=1.0\textwidth]{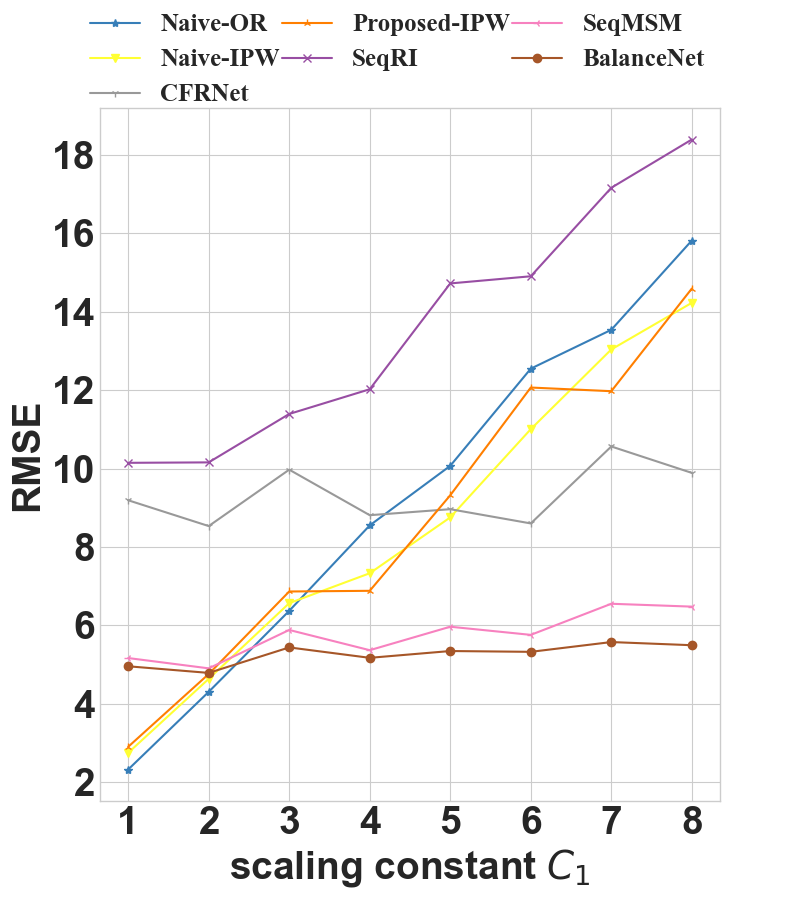}
    \end{minipage}%
    }
    }
          \caption*{Figure 2: Effects of different scaling constant on $\epsilon_{CATE}$.}   \label{fig.3}
  \end{minipage}
  \begin{minipage}[p]{0.35\textwidth}
    \centering
    \caption{Performance metrics under various $(\lambda_1, \lambda_2)$ pairs.}\label{tab:lambda_results}
    \resizebox{\linewidth}{!}{
\begin{tabular}{c|c|c||c|c|c}
\toprule
\multicolumn{3}{c||}{Fixed $\lambda_1=0.2$} & \multicolumn{3}{c}{Fixed $\lambda_2=0.2$} \\
\hline
$\lambda_2$ & $\epsilon_{\text{CATE}}$ & $\epsilon_{\text{ATE}}$ & $\lambda_1$ & $\epsilon_{\text{CATE}}$ & $\epsilon_{\text{ATE}}$ \\
\hline
0.0 & 1.29 & 0.20 & 0.0 & 1.29 & 0.22 \\
0.2 & 1.29 & 0.20 & 0.2 & 1.28 & 0.20 \\
0.5 & 1.29 & 0.23 & 0.5 & 1.29 & 0.19 \\
1.0 & 1.28 & 0.21 & 1.0 & 1.28 & 0.18 \\
1.5 & 1.28 & 0.22 & 1.5 & 1.29 & 0.20 \\
2.0 & 1.29 & 0.22 & 2.0 & 1.29 & 0.21 \\
3.0 & 1.29 & 0.23 & 3.0 & 1.28 & 0.21 \\
4.0 & 1.28 & 0.22 & 4.0 & 1.28 & 0.21 \\
5.0 & 1.28 & 0.23 & 5.0 & 1.28 & 0.18 \\
\bottomrule
\end{tabular}}
  \end{minipage}
\end{table}

\subsection{Results}  
\textbf{Performance comparison.} 
We average the results over 20 trials of estimating the long-term treatment effects in JOBS and IHDP, and the results are presented in Table~\ref{tab:2}. We fix the imbalance penalties $\lambda_1=\lambda_2=1$ and the constant scaling the effect of historical outcomes $C_1=5$, $C_2=2$. The best results are presented in the bold font. With the results shown in Table~\ref{tab:2}, we make the following observations: (1) The proposed methods SeqMSM and BalancedNet outperform the others. This demonstrates the necessity of modeling the dependencies between short-term and long-term outcomes and balancing the missing events and non-missing events. (2) Besides, the BalancedNet outperforms SeqMSM with lower variance (see Appendix~\ref{Appendix:d}), demonstrating its ability to reduce SeqMSM's susceptibility to high variance and mitigate the risk of propensity score and selection score model misspecification.
   (3) The baselines, i.e., Naive-OR, Naive-IPW, and CFRNet, show less competitive performance. This is mainly because they ignore the important information embedded
in the sequence of short-term outcomes.
(4) In addition, the SeqRI method performs worse than the SeqMSM and BalancedNet methods due to its failure to account for missing balance and treatment balance, leading to model extrapolation issues.
 (5)  Finally, the Proposed-IPW method relies solely on non-missing data, neglecting information from the joint missing data and thus resulting in low data utilization efficiency, leading to worse performance.

\textbf{Effects of varying missing ratios.} We study the impact of missing ratios $\gamma \in [0.05,0.5]$ on long-term treatment effects estimation. As shown in Table~\ref{tab:2},  BalanceNet consistently outperforms other methods across nearly all scenarios. In addition, as the missing ratio
increases, the performance of the proposed methods gradually declines. This is expected, as higher
missing ratios mean more long-term outcomes are unobserved, making the task more challenging.

\textbf{Effects of varying scaling constants $C_1$ and $C_2$.} We further study the effects of varying scaling constants $C_1$ and $C_2$. As $C_1$ and $C_2$ increase, the impact of prior outcomes on subsequent outcomes increases. 
In Figure 2, we find that as $C_1$ and $C_2$ increase, our methods achieve the optimal performance in nearly all scenarios. As for the other methods, i.e., Naive-OR, Naive-IPW, Proposed-IPW, and CFRNet, they ignore the important information embedded
in the sequence of short-term outcomes, leading to worse performance as scaling constants increase.

\textbf{Effects of varying imbalance penalties.} In Table \ref{tab:2}, we just set $\lambda_1 = \lambda_2 = 1$.
 In fact, from  Eq.~\eqref{eq:4}, $\lambda_1$ and $\lambda_2$ are two key hyperparameters for the BalanceNet method. 
Thus, we examine the robustness of the BalanceNet method across various $(\lambda_1, \lambda_2)$ pairs. As shown in Table \ref{tab:lambda_results}, the BalanceNet method performs stably,  demonstrating its robustness with respect to $(\lambda_1, \lambda_2)$. 

We also evaluate the proposed methods at different time steps, the results are similar to that of 
Table \ref{tab:2} and are provided in Appendix~\ref{Appendix:e}.
This further demonstrates the robustness of the proposed methods. 

\section{Conclusion}
In this paper, we propose principled approaches to estimate  long-term treatment effects in the setting of multi-stage 
 and monotone missing. First, we introduce a novel sequential missing mechanism assumption that is weaker and better aligns with real-world scenarios, compared with the commonly used missing at random assumption. Second, we propose two basic methods, IPW and SeqRI, to address the missing data issue. By analyzing their limitations, we propose the SeqMSM method, which integrates the advantages of the IPW and SeqRI methods and delivers improved performance. Finally, considering the SeqMSM may suffer from high variance problems, we further 
propose a novel balancing-enhanced neural network-based approach, BalanceNet, to improve the stability of the SeqMSM method for long-term treatment effect estimation. Extensive experiments are conducted on two datasets, and the results show that our methods have the best performance. 

\bibliographystyle{unsrt}
\bibliography{ref}


\appendix

\setcounter{theorem}{0}
\setcounter{equation}{0}
\setcounter{table}{0}
 \renewcommand{\theequation}{A.\arabic{equation}}
  \renewcommand{\thetable}{A\arabic{table}}
    \renewcommand{\thefigure}{A\arabic{figure}}
\newtheorem{lem}{Lemma}[section]
\newtheorem{thm}{Theorem}[section]
\newtheorem{prop}{Proposition}[section]

\newpage 
\section{Related Work}
\textbf{Long-term treatment effect estimation.}  To estimate the long-term causal effect, a common approach is to analyze treatment effects on short-term surrogate outcomes as a proxy for long-term outcomes~\cite{yin2020novel}. This method relies on the stringent surrogacy assumption. This strategy assumes that short-term outcomes are closely related to long-term effects, allowing researchers to draw inferences about long-term treatment effects from more readily available short-term data. However, selecting an appropriate surrogate can be challenging, as it may suffer from the surrogate paradox~\cite{chen2007criteria}. Several recent work seeking to address these issues, such as Athey et al.~\cite{athey2019surrogate} proposed using a set of intermediate outcomes to infer the primary outcome;  Cheng et al.~\cite{cheng2021long} proposed to build sequential models to learn surrogate representations.
However, these works disregard unobserved surrogates and treat all short-term outcomes as surrogates. To address this issue, Cai et al.~\cite{cai-etal2024} developed a flexible method called LASER to estimate long-term causal effects in a situation where the surrogates are either observed or have observed proxies. 
Some studies have investigated the identification and estimation of long-term treatment effects by combining short-term experimental and long-term observational data, such as \cite{kallus2024role, Ghassami-etal2022,  Imbens-etal2024, hu2025-longterm, Goffrier-etal2023}. 
In addition, Tran et al.~\cite{Tran-etal2024}
 proposed method for inferring the long-term causal effects of long-term treatments from short-term experiments in a dynamic setting.  


Monotone missing is a typical problem faced in the estimation of long-term treatment effects. 
In our setting, monotone missing data introduces several significant challenges, such as missing not at random (MNAR) mechanisms and severe data sparsity.  
However, the issue of monotone missing in estimating long-term treatment effects has been rarely discussed. In this paper, we explore this challenging monotone missing data setting for estimating multi-stage long-term treatment effects, which has a wide range of potential applications.

\textbf{Missing data.}
In multi-stage long-term treatment effect estimation, the outcomes often suffer from the issues of missing data. There are three missing data
mechanisms: missing at random (MAR), missing completely at random
(MCAR), and missing not at random (MNAR). Most recent studies addressing missing data typically assume a missing at random (MAR) mechanism, where the probability of missingness is assumed to depend solely on the observed variables~\cite{cheng2021long, Hernan-Robins2020, zhao2024covariate}. 
Monotone missing refers to a missing data pattern where the missingness increases monotonically over time or stages~\cite{molenberghs2014handbook, Li-etal2014}. Several studies have explored monotone missing under the MAR assumption in longitudinal study, such as \cite{Li-etal2014, Robins1986, Robins-etal2000}.  

In estimating long-term treatment effects, unlike previous studies that assume the long-term outcome is MAR, we consider a prevalent and significant form of missingness--monotone missing--under the MNAR assumption (see discussion below Assumption \ref{assum:4-1} in the main text), and propose several novel approaches for effective estimation of long-term treatment effects. 



\section{Data Structure}  \label{app-tmp}

When $T = 3$, the observed data structure is presented in Table \ref{tab:1-app} below.

\begin{table}[ht]
    \centering 
    \caption{Observed data, where \checkmark and $\text{NA}$ mean observed and missing, respectively.} 
    \begin{tabular}{ccccccccc} 
        \toprule 
        Unit & $X$ & $A$ &$R_1$ &$S_1$ &$R_2$ &$S_2$ &$R_3$ &$Y$ \\
        \midrule
        1 & \checkmark & \checkmark &1 & \checkmark & 1 & \checkmark& 1 & \checkmark \\
        $\cdots$ & \checkmark & \checkmark &1 & \checkmark & 1 & \checkmark& 1 & \checkmark \\
        $n_1$ & \checkmark & \checkmark &1 & \checkmark & 1 & \checkmark& 1 & \checkmark \\
        \midrule
        $n_1+1$ & \checkmark & \checkmark &1 & \checkmark & 1 & \checkmark& 0 & $\text{NA}$ \\
        $\cdots$ & \checkmark & \checkmark &1 & \checkmark & 1 & \checkmark & 0 & $\text{NA}$ \\
        $n_2$ & \checkmark & \checkmark &1 & \checkmark & 1 & \checkmark & 0 & $\text{NA}$ \\
        \midrule
        $n_2+1$ & \checkmark & \checkmark &1 & \checkmark & 0 & $\text{NA}$& 0 & $\text{NA}$ \\
        $\cdots$ & \checkmark & \checkmark &1 & \checkmark & 0 & $\text{NA}$ & 0 & $\text{NA}$ \\
        $n_3$ & \checkmark & \checkmark &1 & \checkmark & 0 & $\text{NA}$ & 0 & $\text{NA}$ \\
        \midrule
        $n_3+1$ & \checkmark & \checkmark &0 & $\text{NA}$ & 0 & $\text{NA}$ & 0 & $\text{NA}$ \\
        $\cdots$ & \checkmark & \checkmark &0 & $\text{NA}$ & 0 & $\text{NA}$ & 0 & $\text{NA}$ \\
        $n$ & \checkmark & \checkmark &0 & $\text{NA}$ & 0 & $\text{NA}$ & 0 & $\text{NA}$ \\
        \bottomrule 
    \end{tabular}
    \label{tab:1-app} 
\end{table}

\section{Technical Proofs} \label{proofs-app}  

\textbf{Theorem \ref{them:4-2}} (Identifiability).
Under Assumptions \ref{assum:2-1}, \ref{assum:2-2}, and \ref{assum:4-1}, $\tau$ and $\tau(x)$ are identifiable.

\emph{\bf Proof of Theorem \ref{them:4-2}}. 
For $\tau(x)$, we first note that 
\begin{equation*}
    \begin{aligned}
        \tau&(x) = \mathbb{E}[Y(1)-Y(0)\mid X=x]\\
        &=\mathbb{E}\Big [\mathbb{E}  [Y(1)\mid X=x,S_1(1),S_2(1)] ~\Big|~ X=x \Big ]-\mathbb{E} \Big [\mathbb{E}[Y(0)\mid X=x,S_1(0),S_2(0)] ~\Big|~ X=x \Big ]\\
        &=\mathbb{E} \Big [\mathbb{E}[Y(1)\mid X=x,S_1(1),S_2(1),A=1] ~\Big |~ X=x \Big ] -\mathbb{E}\Big [\mathbb{E}[Y(0)\mid X=x,S_1(0),S_2(0),A=0] ~\Big |~ X=x\Big ]\\
        &=\mathbb{E}\Big[\mathbb{E}[Y\mid X=x,S_1,S_2,A=1,R_3=1] ~\Big |~ X=x\Big]-\mathbb{E}\Big[\mathbb{E}[Y\mid X=x,S_1,S_2,A=0,R_3=1]~\Big |~ X=x \Big ],
    \end{aligned}
\end{equation*}
 where the second equality follows by the law of iterated expectations, the third equality follows from Assumption~\ref{assum:2-2}, and
 the fourth equality follows from Assumption~\ref{assum:4-1}.
Then the identifiability of $\tau(x)$ can be obtained by noting that both $\mathbb{E}[Y\mid X=x,S_1,S_2,A=1,R_3=1]$ and $\mathbb{E}[Y\mid X=x,S_1,S_2,A=0,R_3=1]$ are identifiable by Assumption~\ref{assum:2-1}. 

Likewise, the identifiability of $\tau$ can be obtained by
\begin{equation*}
    \begin{aligned}
        \tau&=\mathbb{E}[\mathbb{E}[Y(1)-Y(0)\mid X=x]]\\
        &=\mathbb{E}\{\mathbb{E}[\mathbb{E}[Y\mid X=x,S_1,S_2,A=1,R_3=1]\mid X=x] \\
        &\quad -\mathbb{E}[\mathbb{E}[Y\mid X=x,S_1,S_2,A=0,R_3=1]\mid X=x]\}.
    \end{aligned}
\end{equation*}
This finishes the proof. 

\hfill $\Box$


\bigskip 
We prove Eq. \eqref{eq2} in the main text, that is, 
 \begin{equation} \tag{2}
  \mu_a(x) = \mathbb{E} \left [ \frac{\mathbb{I}(A=a) R_3 Y}{ e_a(X) r_3(X, A, S_1, S_2) } \Big | X=x  \right],a=0,1,
\end{equation} 
where $e_a(X)=\mathbb{P}(A=a \mid X)$ is the propensity score and  $r_3(X,A,S_1,S_{2})=\mathbb{P}(R_3=1\mid X,A,S_1,S_{2})$ is the selection score.

\bigskip 
\emph{\bf Proof of Eq. \eqref{eq2}}. Eq. \eqref{eq2} holds immediately by noting that    
  \begin{align*}
        & \mathbb{E} \left [ \frac{\mathbb{I}(A=a) R_3 Y}{ e_a(X) r_3(X, A, S_1, S_2) } \Big | X=x  \right] \\
       ={}&  \mathbb{E} \left [ \frac{\mathbb{I}(A=a) R_3 Y(a)}{ e_a(X) r_3(X, A=a, S_1(a), S_2(a)) } \Big | X=x  \right]  \\
          ={}&  \mathbb{E} \left [ \frac{\mathbb{I}(A=a) R_3 Y(a)}{ e_a(X) r_3(X, A=a, S_1(a), S_2(a)) } \Big | X=x, A=a  \right] \mathbb{P}(A=a\mid X)  \\
             ={}&  \mathbb{E} \left [ \frac{ R_3 Y(a)}{  r_3(X, A=a, S_1(a), S_2(a)) } \Big | X=x, A=a  \right]  \\
       ={}& \mathbb{E} \left [ \mathbb{E} \left \{  \frac{ R_3 Y(a)}{ r_3(X, A=a, S_1(a), S_2(a)) } \Big | X=x, A=a, S_1(a), S_2(a) \right \} \Big | X=x, A=a  \right] \\
       ={}& \mathbb{E} \left [ \mathbb{E} \left \{  \frac{ \mathbb{P}(R_3=1\mid X=x, A=a, S_1(a), S_2(a) ) \cdot \mathbb{E}[Y(a)\mid X=x, A=a, S_1(a), S_2(a) ] }{ r_3(X, A=a, S_1(a), S_2(a)) }  \right \} \Big | X=x, A=a  \right] \\
       ={}& \mathbb{E} \left [  \mathbb{E}[Y(a)\mid X=x, A=a, S_1(a), S_2(a) ]  \Big | X=x, A=a  \right] \\
       ={}& \mathbb{E}(Y(a)\mid X=x, A=a) \\
       ={}& \mathbb{E}(Y(a)\mid X=x) \\
       ={}& \mu_a(x),
    \end{align*}
 where the fourth equality follows from the law of iterated expectations, the fifth equality from Assumption \ref{assum:4-1}, and the eighth equality from Assumption \ref{assum:2-2}. 
 
\hfill $\Box$

\section{Experimental Details in the Main Text}
\label{Appendix:c}

\subsection{Datasets}

The IHDP dataset is based on the Infant Health and Development Program (IHDP) and investigates the effectiveness of high-quality home visits in promoting children's future cognitive development. The dataset comprises 747 units, including 139 treated and 608 controlled, and has 25 features that provide a comprehensive picture of the children and their mothers. The JOBS dataset is based on the National Supported Work program and examines the effects of job training on income and employment status. The dataset consists of 2,570 
units (237 treated, 2,333 controlled), with 17 features. For both datasets, we use the original features, treatment
assignment, and synthetic outcomes (more details later) that reflect the sequential observations in long-term treatment estimation.

\subsection{Simulating Outcomes}
Following the previous studies~\cite{wu2024policy, cheng2021long}, we assume that the outcome observed at timestep $t$ is determined by three factors: features $X_i$, treatment $A_i$ and
all the outcomes observed before timestep $t$.

For the IHDP dataset, since the short-term outcome $S_1$ is only determined by the features $X_i$ and treatment $A_i$, the potential outcome of unit $i$ at $t$ is generated as follows:
\begin{equation}\label{eq:5}
S_{1,i}(a)\sim
\begin{cases}
    \textup{b}(\sigma(w_0X_i+\epsilon_{0,i})),&a=0 \\
    \textup{b}(\sigma(w_1X_i+\epsilon_{1,i})),&a=1
\end{cases}
\end{equation}
where ``b'' denotes ``bernoulli distribution'', $\sigma(\cdot)$ is the sigmoid function, $w_0\sim\mathcal{N}_{[-1,1]}(0,1)$ follows a truncated normal distribution, $w_1\sim\textup{Unif}(-1,1)$ follows a uniform distribution, $\epsilon_{0,i}\sim\mathcal{N}(\mu_0,\sigma_0)$, $\epsilon_{1,i}\sim\mathcal{N}(\mu_1,\sigma_1)$, and set $\mu_0=1,\mu_1=3$ and $\sigma_0=\sigma_1=1$ for the IHDP dataset.
The short-term outcome $S_t$ ($t\in\{2,...,t_0\}$) and the long-term outcome $Y$ are generated as follows:
\begin{align*}
S_{t,i}(a)&\sim
\begin{cases}
    \mathcal{N}(\beta_0X_i,1)+C_1\sum\nolimits_{j=1}^{t-1}S_{j,i}(0),a=0 \\
    \mathcal{N}(\beta_1X_i+2,0.5)+C_1\sum\nolimits_{j=1}^{t-1}S_{j,i}(1),a=1
\end{cases}     \\   
Y_i(a)&\sim
\begin{cases}
    \mathcal{N}(\beta_0X_i,1)+C_1\sum\nolimits_{j=1}^{t_{0}}S_{j,i}(0),a=0 \\
    \mathcal{N}(\beta_1X_i+2,0.5)+C_1\sum\nolimits_{j=1}^{t_{0}}S_{j,i}(1), a=1
\end{cases}
\end{align*}
where $\beta_0$ is randomly sampled from $\{0,1,2,3,4\}$ with probabilities $\{0.5,0.2,0.15,0.1,0.05\}$, $\beta_1\sim4\cdot\mathcal{N}_{[0,4]}(0,1)$, and $C_1$ is a constant scaling the effect of historical outcomes.

Similarly, for the JOBS dataset, we employ Eq.~\eqref{eq:5} to generate $S_1$ and set $\mu_0 = 0, \mu_1 = 2$ and $\sigma_0 = \sigma_1 = 1$.
In addition, we generate $S_t$ ($t\in\{2,...,t_0\}$) and $Y$ as follows: 
\begin{align*}
&S_{t,i}(a)\sim
\begin{cases}
    \textup{b}(\sigma(\beta_0X_i)+\frac{C_2}{t-1}\sum_{j=1}^{t-1}S_{j,i}(0))+\epsilon_{0,i},a=0 \\
    \textup{b}(\sigma(\beta_1X_i)+\frac{C_2}{t-1}\sum_{j=1}^{t-1}S_{j,i}(1))+\epsilon_{1,i},a=1
\end{cases} \\
&Y_i(a)\sim
\begin{cases}
    \textup{b}(\sigma(\beta_0X_i))+\frac{C_2}{t_0}\sum_{j=1}^{t_0}S_{j,i}(0)+\epsilon_{0,i},a=0 \\
    \textup{b}(\sigma(\beta_1X_i))+\frac{C_2}{t_0}\sum_{j=1}^{t_0}S_{j,i}(1)+\epsilon_{1,i},a=1
\end{cases}
\end{align*}
where $\epsilon_{0,i}\sim\mathcal{N}(0,1)$, $\epsilon_{1,i}\sim\mathcal{N}(0,0.5)$, and $C_2$ is a constant scaling the effect of historical outcomes.

We set $t_0 = 2$ or $T=3$ for both simulated datasets in the main text. In Appendix \ref{Appendix:e}, we conduct experiments for case of $T=4$.

\subsection{Missing Mechanism}
Following the missing mechanism in Assumption~\ref {assum:2-1} and Assumption~\ref{assum:4-1}. we specify a missing ratio $\gamma$.
Eventually, taking $T=3$ as an example, we adopt the missing mechanism below to generate the missing indicator $\boldsymbol{R}$. 
For the $S_1$, the short-term outcome $S_1$ is missing at random, we randomly select $\gamma N$ units to be missing. For the $S_2$, we first select the unit $i$ with $R_{1, i}=0$ to be missing, and then calculate the value of $\textup{score}_i$ $=S_{1, i}+\sum_{j=1}^p X_{ij}$ from the remaining $(1-\gamma) N$ units with $R_{1,i}=1$. Then, we select the $\gamma (1-\gamma) N$ units with the smallest $\textup{score}_i$ to be missing. For the long-term outcome $Y$, similarly, we first select the unit $i$ with $R_{2, i}=0$ to be missing, and then calculate the value of $\textup{score}_i$ $=S_{1, i}+S_{2, i}+\sum_{j=1}^p X_{ij}$ from the remaining $(1-\gamma)^2 N$ units with $R_{2,i}=1$. Then, we select the $\gamma (1-\gamma)^2 N$ units with the smallest $\textup{score}_i$ to be missing.

\subsection{Implementation Details}

The outcome regression is modeled using a Multi-Layer Perceptron (MLP), while both the propensity score and selection score are estimated via logistic regression.  
 The table below presents the hyperparameter space for each method in both IHDP and JOBS datasets.  

\begin{table*}[h]
\renewcommand \arraystretch{1.5} 
\centering
\caption{Implementation Details}
\resizebox{1 \linewidth}{!}{
\begin{tabular}{l|c|c|c|c}
\toprule
\textbf{Hyperparameter} & \textbf{Naive-OR} & \textbf{Naive-IPW} & \textbf{CFRNet} & \textbf{Proposed-IPW}  \\
\hline
Learning rate & \multirow{5}{*}{\begin{tabular}[c]{@{}c@{}}MLPRegressor\\(hidden\_layer\_sizes=(50,25),\\ max\_iter=100000, random\_state=42)\end{tabular}} & [0.001,0.005,0.01] & [0.001,0.005,0.01] & [0.001,0.005,0.01]  \\ 
Batch size & & Full Batch & Full Batch & Full Batch  \\
Architecture & & [1-3] hidden layers & [1-3] hidden layers & [1-3] hidden layers  \\
Optimizer & & Adam & Adam & Adam  \\
Early stopping patience & & 10 & 10 & 10  \\
Activation function (all layers) & & ReLU & ELU & ReLU  \\
\hline
\textbf{Hyperparameter} &   \textbf{SeqRI} & \textbf{SeqMSM} & \textbf{BalanceNet} \\
\hline   
Learning rate &  \multirow{5}{*}{\begin{tabular}[c]{@{}c@{}}MLPRegressor\\(hidden\_layer\_sizes=(50,25),\\ max\_iter=100000, random\_state=42)\end{tabular}} & [0.001,0.005,0.01] & [0.001,0.005,0.01] \\ 
Batch size & &  Full Batch & Full Batch \\
Architecture  & & [1-3] hidden layers & [1-3] hidden layers \\
Optimizer  & & Adam & Adam \\
Early stopping patience &  & 10 & 10 \\
Activation function (all layers) &  & ReLU & ELU \\
\bottomrule
\end{tabular}}
\end{table*}

From this table above, one can see that we present a fair comparison of various methods.

\section{Additional Experimental Results}\label{Appendix:d}

We provide additional results for the standard deviation of different methods in Table~\ref{tab3-appendix}, the effect of different imbalance penalties in Table~\ref{tab4-appendix}.


\begin{table*}[h]
\renewcommand \arraystretch{1.25} 
\centering
\caption{Performance comparison of variance on IHDP dataset and JOBS dataset with $\lambda_1=\lambda_2=1,C_1=5,C_2=2$.}
\resizebox{1 \linewidth}{!}{
\begin{tabular}{c|cc |cc|cc|cc|cc|cc|cc}
\toprule
\multicolumn{1}{c|}{JOBS} & \multicolumn{2}{c|}{$\gamma=0.05$}  & \multicolumn{2}{c|}{$\gamma=0.1$}   & \multicolumn{2}{c|}{$\gamma=0.15$}   
& \multicolumn{2}{c|}{$\gamma=0.2$}        & \multicolumn{2}{c|}{$\gamma=0.3$}    
& \multicolumn{2}{c|}{$\gamma=0.4$}
& \multicolumn{2}{c}{$\gamma=0.5$}
\\ \midrule
 Method & $\epsilon_{{CATE}_{std}}$  & $\epsilon_{{ATE}_{std}}$ & $\epsilon_{{CATE}_{std}}$  & $\epsilon_{{ATE}_{std}}$ & $\epsilon_{{CATE}_{std}}$  & $\epsilon_{{ATE}_{std}}$  & $\epsilon_{{CATE}_{std}}$  & $\epsilon_{{ATE}_{std}}$ &$\epsilon_{{CATE}_{std}}$  & $\epsilon_{{ATE}_{std}}$ &$\epsilon_{{CATE}_{std}}$  & $\epsilon_{{ATE}_{std}}$ & $\epsilon_{{CATE}_{std}}$  & $\epsilon_{{ATE}_{std}}$    
 \\ \midrule

Naive-OR &1.798  & 0.769  & 2.393  & 1.190  & 1.212  & 0.863  & 1.472  & 0.912  & 1.529  & 1.040  & 2.198  & 1.029  & 1.076  & 0.858   \\
Naive-IPW &3.258  & 1.829  & 3.639  & 2.366  & 3.288  & 1.733  & 3.664  & 1.742  & 2.875  & 1.658  & 2.499  & 1.519  & 1.458  & 0.974   \\
CFRNet &0.278  & 0.458  & 0.331  & 0.507  & 0.298  & 0.492  & 0.334  & 0.499  & 0.329  & 0.460  & 0.467  & 0.619  & 0.406  & 0.571   \\
\midrule
Proposed-IPW &3.657  & 1.959  & 2.360  & 1.466  & 2.932  & 1.530  & 3.683  & 2.049  & 3.929  & 1.850  & 1.916  & 0.992  & 2.341  & 1.370   \\
SeqRI &2.295  & 1.266  & 2.885  & 1.212  & 1.708  & 1.074  & 5.837  & 3.159  & 2.092  & 1.278  & 2.115  & 1.371  & 1.763  & 1.061   \\
SeqMSM &0.108  & 0.220  & 0.132  & 0.355  & 0.177  & 0.336  & 0.186  & 0.319  & 0.280  & 0.478  & 0.254  & 0.389  & 0.242  & 0.430   \\
BalanceNet & \textbf{0.048}  & \textbf{0.145}  & \textbf{0.053}  & \textbf{0.168}  & \textbf{0.061}  & \textbf{0.170}  & \textbf{0.070}  & \textbf{0.182}  & \textbf{0.055}  & \textbf{0.162}  & \textbf{0.049}  & \textbf{0.163}  & \textbf{0.064}  & \textbf{0.189}   \\

  \midrule

  \midrule
  \multicolumn{1}{c|}{IHDP}   & \multicolumn{2}{c|}{$\gamma=0.05$}  & \multicolumn{2}{c|}{$\gamma=0.1$}   & \multicolumn{2}{c|}{$\gamma=0.15$}   
& \multicolumn{2}{c|}{$\gamma=0.2$}        & \multicolumn{2}{c|}{$\gamma=0.3$}    
& \multicolumn{2}{c|}{$\gamma=0.4$}
& \multicolumn{2}{c}{$\gamma=0.5$}
\\ \midrule
 Method & $\epsilon_{{CATE}_{std}}$  & $\epsilon_{{ATE}_{std}}$ & $\epsilon_{{CATE}_{std}}$  & $\epsilon_{{ATE}_{std}}$ & $\epsilon_{{CATE}_{std}}$  & $\epsilon_{{ATE}_{std}}$  & $\epsilon_{{CATE}_{std}}$  & $\epsilon_{{ATE}_{std}}$ &$\epsilon_{{CATE}_{std}}$  & $\epsilon_{{ATE}_{std}}$ &$\epsilon_{{CATE}_{std}}$  & $\epsilon_{{ATE}_{std}}$ & $\epsilon_{{CATE}_{std}}$  & $\epsilon_{{ATE}_{std}}$    
 \\ \midrule 
Naive-OR &\textbf{0.780}  & 0.643  & 1.412  & 1.501  & 1.423  & 1.375  & 2.309  & 2.084  & 7.436  & 4.823  & 3.494  & 3.135  & 2.302  & 2.819   \\
Naive-IPW &3.468  & 2.709  & 5.665  & 3.772  & 2.164  & 1.282  & 4.029  & 2.609  & 4.864  & 2.790  & 5.570  & 4.529  & 5.837  & 4.012   \\
CFRNet &2.927  & 1.609  & 2.894  & 1.531  & 3.987  & 2.123  & 3.070  & 1.979  & 3.076  & 1.541  & 4.818  & 2.305  & 5.250  & 2.883   \\
\midrule
Proposed-IPW &2.864  & 2.077  & 2.151  & 1.137  & 4.214  & 2.944  & 3.423  & 2.504  & 4.221  & 3.072  & 6.284  & 4.964  & 5.925  & 4.066   \\
SeqRI &1.760  & 1.507  & 3.090  & 2.281  & 3.547  & 1.594  & 4.901  & 2.694  & 4.221  & 2.827  & 3.373  & 3.357  & 2.877  & 2.477   \\
SeqMSM &1.622  & 0.592  & 1.516  & 0.692  & 1.439  & 0.635  & 1.706  & 1.086  & 1.413  & 1.408  & 1.310  & 1.199  & 2.088  & 2.366   \\
BalanceNet&1.212  & \textbf{0.117}  & \textbf{1.222}  & \textbf{0.114}  & \textbf{1.233}  & \textbf{0.100}  & \textbf{1.270}  & \textbf{0.142}  & \textbf{1.156}  & \textbf{0.112}  & \textbf{1.147}  & \textbf{0.140}  & \textbf{1.197}  & \textbf{0.117}   \\

  \bottomrule
  \end{tabular}}
  \label{tab3-appendix}
\end{table*}%

\bigskip

\begin{table*}[h]
\renewcommand \arraystretch{1.25} 
\centering 
\caption{Performance comparison under different imbalance penalty $\lambda_1=\lambda_2=\lambda$ on IHDP dataset and JOBS dataset.}
\resizebox{1 \linewidth}{!}{
\begin{tabular}{c|cc|cc|c|cc|cc}
\toprule
\multicolumn{1}{c|}{JOBS}  & \multicolumn{2}{c|}{CFRNet}   & \multicolumn{2}{c|}{BalanceNet} &\multicolumn{1}{c|}{IHDP} & \multicolumn{2}{c|}{CFRNet}   & \multicolumn{2}{c}{BalanceNet}
\\ \midrule
 Imbalance Penalty ($\lambda$) & $\epsilon_{{CATE}}$  & $\epsilon_{{ATE}}$ & $\epsilon_{{CATE}}$  & $\epsilon_{{ATE}}$ & Imbalance Penalty ($\lambda$) & $\epsilon_{{CATE}}$  & $\epsilon_{{ATE}}$ & $\epsilon_{{CATE}}$  & $\epsilon_{{ATE}}$    
 \\ \midrule
$\lambda=0.2$ &2.036  & 1.093  & \textbf{1.292}  & \textbf{0.215} &$\lambda=0.2$ &8.964 & \textbf{1.827} & \textbf{5.378} & 2.045 \\
$\lambda=0.5$ &2.031  & 1.175  & \textbf{1.282}  & \textbf{0.186}  &$\lambda=0.5$ &8.964 & \textbf{1.827} & \textbf{5.378} & 2.045  \\
$\lambda=1.0$ & 1.882  & 1.100  & \textbf{1.293}  & \textbf{0.220}  &$\lambda=1.0$ &8.799 & \textbf{1.908} & \textbf{5.373} & 2.022   \\
$\lambda=1.5$&1.912  & 1.173  & \textbf{1.288}  & \textbf{0.222}  &$\lambda=1.5$&9.880 & \textbf{1.853} & \textbf{5.363} & 2.024  \\
$\lambda=2.0$&1.892  & 1.147  & \textbf{1.296}  & \textbf{0.247}  & $\lambda=2.0$ &9.278 & 2.454 & \textbf{5.347} & \textbf{1.970}  \\
$\lambda=3.0$ &1.893  & 1.191  & \textbf{1.291}  & \textbf{0.227} & $\lambda=3.0$ &9.028 & 1.979 & \textbf{5.306} & \textbf{1.965}\\
$\lambda=4.0$ &1.868  & 1.168  & \textbf{1.286}  & \textbf{0.216} & $\lambda=4.0$ &9.278 & 2.298 & \textbf{5.308} & \textbf{1.973}\\
$\lambda=5.0$ &1.846  & 1.153  & \textbf{1.284}  & \textbf{0.219} &$\lambda=5.0$ &9.109 & 2.282 & \textbf{5.277} & \textbf{1.943}\\

  \bottomrule
  \end{tabular}}
  \label{tab4-appendix}
\end{table*}%

As shown in Table~\ref{tab3-appendix}, the BalancedNet outperforms SeqMSM with lower variance, demonstrating its ability to reduce SeqMSM's susceptibility to high variance. In Table~\ref{tab4-appendix}, we can find that
BalanceNet consistently outperforms other methods in almost all imbalance penalties.

\newpage 
\section{Extension to $T > 3$}   \label{Appendix:add}

Our proposed methods, including IPW, SeqMSM and BalanceNet, are not limited to $T=3$ and can be naturally extended to scenarios where $T>3$. In this section, we provide a detailed description of this extension.

\subsection{IPW Method}
For the IPW method (Section 4.2 of the main text), 
the key quantity for identifiability is the selection score, given by 
 $$r_T(X, A, S_1, S_2, ..., S_{t_0}) = \mathbb{P}(R_T = 1| X,A, S_1, S_2, ..., S_{t_0}),$$
where $t_0 = T-1$.  
When $T > 3$, $r_T(X, A, S_1, S_2, ..., S_{t_0})$ remains identifiable using the same analysis as in Proposition \ref{prop:4.3}. Moreover, by a proof analogous to that of Eq.~\eqref{eq2}, we can show that 
     \begin{equation} \label{eq-A2}
  \mu_a(x) = \mathbb{E} \left [ \frac{\mathbb{I}(A=a) R_T Y}{ e_a(X) r_T(X, A, S_1, S_2, \cdots, S_{t_0}) } \Big | X=x  \right], \quad a=0,1,
\end{equation} 
where $e_a(X)=\mathbb{P}(A=a \mid X)$ is the propensity score. Based on \eqref{eq-A2}, the IPW estimator for $\mu_a(x)$ can be obtained by regressing $\frac{\mathbb{I}(A=a) R_T Y}{ e_a(X) \hat r_T(X, A, S_1, S_2, \cdots, S_{t_0}) } $ on $X$, where $\hat r_T(X, A, S_1, S_2, ..., S_{t_0})$ is the estimate of $r_T(X, A, S_1, S_2, ..., S_{t_0})$.

\subsection{Sequential Regression Imputation Method}

Note that under Assumptions~\ref{assum:2-2} and~\ref{assum:4-1},  
\begin{align*}
       \mathbb{E}[ S_1 | X, A=a, R_1 = 0 ] ={}& \mathbb{E}[ S_1 | X, A=a, R_1 = 1 ] \\
       \mathbb{E}[ S_2 |  X, A=a, S_1, R_2 = 0 ] ={}& \mathbb{E}[ S_2 | X, A=a, S_1, R_2 = 1 ] \\
       \vdots \\ 
       \mathbb{E}[ Y | X, A=a, S_1, S_2, \cdots, S_{t_0}, R_t = 0 ] ={}& \mathbb{E}[ Y | X, A=a, \cdots, S_{t_0}, R_t = 1 ].    \end{align*}
 Thus, due to the monotone missing pattern,  we can impute the outcomes $S_1, S_2, \cdots, S_{t_0}$ and $Y$ sequentially by modeling $m_{1a}(X) := \mathbb{E}[S_1 | X, A=a, R_1 = 1 ]$, $m_{2a}(X, S_1) := \mathbb{E}[ S_2 | X, A=a, S_1, R_2 = 1 ]$, $\cdots$, 
 and $m_{Ta}(X, S_1, S_2, \cdots, S_{t_0}) := \mathbb{E}[ Y | X, A=a, S_1, S_2, \cdots, S_{t_0} R_T = 1 ]$, respectively. Let $\hat m_{1a}, \hat m_{2a}, \cdots,  \hat m_{Ta}$ denote the estimates of $m_{1a},m_{2a}, \cdots,$ and $m_{Ta}$, respectively. The corresponding estimation procedures are summarized in Algorithm~\ref{alg:2}. 

\begin{algorithm}
\caption{Sequential regression imputation algorithm}
\label{alg:2}
\begin{algorithmic}[1]
\STATE Imputing the missing value of $S_1(a)$ using $\hat{m}_{1a}(X)$ and regard  $S_1(a)$ as ``non-missing".
\STATE Imputing the missing value of $S_2(a)$ using $\hat{m}_{2a}(X,S_1)$ with $(X,\hat{m}_{1a}(X))$ as the input. Then we regard $S_2(a)$ as ``non-missing" for $a=0,1$.

\STATE $\cdots$

\STATE Predicting $Y(a)$ using $\hat{m}_{Ta}(X,S_{1},S_{2}, \cdots, S_{t_0})$ with $X,\hat{m}_{1a}(X),\hat{m}_{2a}(X,S_1), \cdots,$ and $\hat{m}_{t_0a}(X,S_1, S_2, \cdots, S_{t_0-1})$ as the input. 
\end{algorithmic}
\end{algorithm}

Then, we can estimate $\tau(x)$ with  
$
    \hat{\tau}_{seqri}(x)=\hat{\mu}_1^{seqri}(x)-\hat{\mu}_0^{seqri}(x),
$
where 
$\hat{\mu}_a^{seqri}(x)=\hat{m}_{Ta}(x,\hat{m}_{1a}(x),\hat{m}_{2a}(x,s_1), \cdots, \hat{m}_{t_0a}(x,s_{t_0-1}))$.

\subsection{SeqMSM Method}

For the SeqMSM method (Section 4.4 of the main text), 
the estimation procedure presented in Section 4.4.2 extends naturally to $T>3$ by sequentially learning $f_1(\cdot), f_2(\cdot), ..., f_T(\cdot)$ sequentially. Specifically, the estimation procedures are given as follows.

\emph{Step 1}: Learn $f_1$ by minimizing 
    $
    \mathcal{L}_{1}(f_1)=\sum_a\mathbb{E}_n[\mathbb{I}(A=a)ww_1^{\dagger}\{S_1-f_1(a,X)\}]^2,
    $ 
Denote $\hat f_1$ as the learned $f_1$. 

\emph{Step 2}: Learning $f_2$ by minimizing 
    $
    \mathcal{L}_{2}(f_2)=\sum_a\mathbb{E}_n[\mathbb{I}(A=a)ww_2^{\dagger}\{S_2-f_2(a,X, \hat f_1)\}]^2.
    $
Denote $\hat f_2$ as the learned $f_2$. 

\emph{Step 3:} $\cdots \cdots$ 

\emph{Step 4}: Learn $f_T$ by minimizing 
    $
      \mathcal{L}_{T}(f_T)=\sum_a\mathbb{E}_n[\mathbb{I}(A=a)ww_T^{\dagger}\{Y-f_T(a,X,\hat{f}_1,\hat{f}_2), \cdots, \hat{f}_{t_0}\}]^2,
    $
where $w =1/\hat e_a(X)$, and $w_t^{\dagger}=R_t/\hat{r}_t, t\in\{1,2,3, \cdots, T\}$. 
Denote $\hat f_T$ as the learned $f_T$.

Then, we can estimate $\tau(x)$ with  
$
    \hat{\tau}_{seqmsm}(x)=\hat{\mu}_1^{seqmsm}(x)-\hat{\mu}_0^{seqmsm}(x),
$
where 
$\hat{\mu}_a^{seqmsm}(x)=\hat f_T(a,x,\hat{f}_1,\hat{f}_2, \cdots, \hat{f}_{t_0})$ for $a  = 0, 1$.

\subsection{BalanceNet Method}

For the BalanceNet method (Section 5 of the main text), its sequential structure (Figure~\ref{fig:1}) naturally accommodates cases where $T>3$ by learning balanced representations and predicting missing short-term and long-term outcomes in a sequential manner.

 
\section{Additional Experimental Results When  $T=4$}\label{Appendix:e}

We provide additional results at $T=4$. We
fix the imbalance penalties $\lambda_1=\lambda_2=1$ and the constant scaling the effect of historical outcomes $C_1=3$ and $C_2=2$.

\begin{table*}[h]
\renewcommand \arraystretch{1.25} 
\centering
\caption{Performance comparison on IHDP dataset and JOBS dataset with $\lambda_1=\lambda_2=1,C_1=3,C_2=2$.}
\resizebox{1.05 \linewidth}{!}{
\begin{tabular}{c|cc |cc|cc|cc|cc|cc|cc}
\toprule
\multicolumn{1}{c|}{JOBS} & \multicolumn{2}{c|}{$\gamma=0.05$}  & \multicolumn{2}{c|}{$\gamma=0.1$}   & \multicolumn{2}{c|}{$\gamma=0.15$}   
& \multicolumn{2}{c|}{$\gamma=0.2$}        & \multicolumn{2}{c|}{$\gamma=0.3$}    
& \multicolumn{2}{c|}{$\gamma=0.4$}
& \multicolumn{2}{c}{$\gamma=0.5$}
\\ \midrule
 Method & $\epsilon_{CATE}$  & $\epsilon_{ATE}$ & $\epsilon_{CATE}$  & $\epsilon_{ATE}$ & $\epsilon_{CATE}$  & $\epsilon_{ATE}$ & $\epsilon_{CATE}$ & $\epsilon_{ATE}$   &$\epsilon_{CATE}$ & $\epsilon_{ATE}$  & $\epsilon_{CATE}$  & $\epsilon_{ATE}$  & $\epsilon_{CATE}$    & $\epsilon_{ATE}$    
 \\ \midrule 

Naive-OR &2.584  & 0.927  & 4.672  & 1.396  & 4.419  & 1.189  & 3.791  & 1.374  & 2.967  & 1.403  & 2.710  & 1.420  & 2.696  & 1.543   \\
Naive-IPW &5.932  & 2.330  & 5.788  & 2.110  & 4.989  & 1.586  & 4.525  & 1.510  & 4.128  & 1.280  & 3.308  & 1.132  & 4.042  & 1.747   \\
CFRNet &1.797  & 1.010  & 1.968  & 1.252  & 2.026  & 1.386  & 2.105  & 1.441  & 2.239  & 1.587  & 2.342  & 1.533  & 2.435  & 1.519   \\
\midrule
Proposed-IPW &6.702  & 2.693  & 6.029  & 2.054  & 4.565  & 1.456  & 5.111  & 1.878  & 4.050  & 1.455  & 3.581  & 1.381  & 3.268  & 1.658   \\
SeqRI&2.839  & 0.907  & 5.084  & 1.634  & 4.654  & 1.489  & 4.456  & 1.308  & 3.658  & 1.087  & 3.170  & 0.920  & 3.173  & 1.012   \\
SeqMSM&1.555  & 0.854  & 1.700  & 1.100  & 1.776  & 1.195  & 1.803  & 1.234  & 1.878  & 1.297  & 1.863  & 1.252  & 1.881  & 1.303   \\ 
BalanceNet &\textbf{1.295}  & \textbf{0.272}  & \textbf{1.294}  & \textbf{0.268}  & \textbf{1.296}  & \textbf{0.259}  & \textbf{1.291}  & \textbf{0.254}  & \textbf{1.312}  & \textbf{0.249}  & \textbf{1.308}  & \textbf{0.243}  & \textbf{1.321}  & \textbf{0.282}   \\

  \midrule
  \midrule
  \multicolumn{1}{c|}{IHDP}   & \multicolumn{2}{c|}{$\gamma=0.05$}  & \multicolumn{2}{c|}{$\gamma=0.1$}   & \multicolumn{2}{c|}{$\gamma=0.15$}   
& \multicolumn{2}{c|}{$\gamma=0.2$}        & \multicolumn{2}{c|}{$\gamma=0.3$}    
& \multicolumn{2}{c|}{$\gamma=0.4$}
& \multicolumn{2}{c}{$\gamma=0.5$}
\\ \midrule
 Method &$\epsilon_{CATE}$  & $\epsilon_{ATE}$  &$\epsilon_{CATE}$  & $\epsilon_{ATE}$  &$\epsilon_{CATE}$  & $\epsilon_{ATE}$ & $\epsilon_{CATE}$ & $\epsilon_{ATE}$   &$\epsilon_{CATE}$ & $\epsilon_{ATE}$  & $\epsilon_{CATE}$  & $\epsilon_{ATE}$  & $\epsilon_{CATE}$    & $\epsilon_{ATE}$   
 \\ \midrule 

Naive-OR &11.197  & 2.399  & 12.856  & 3.792  & 13.998  & 5.252  & 15.947  & 6.893  & 19.040  & 8.848  & 20.289  & 10.148  & 19.321  & 10.265   \\
Naive-IPW &9.029  & 2.329  & 10.600  & 3.246  & 14.290  & 6.381  & 16.067  & 7.721  & 19.554  & 8.260  & 18.995  & 8.880  & 17.542  & 9.138   \\
CFRNet &10.236  & \textbf{1.853}  & 9.847  & \textbf{1.317}  & 11.134  & 2.411  & 9.336  & 2.661  & 10.559  & 3.867  & 13.208  & 3.910  & 12.255  & 4.729   \\
\midrule
Proposed-IPW &10.021  & 2.768  & 10.988  & 3.166  & 12.958  & 4.730  & 15.052  & 7.256  & 15.096  & 6.845  & 16.399  & 6.856  & 19.430  & 9.460   \\
SeqRI&14.235  & 3.577  & 15.115  & 2.692  & 15.246  & 2.016  & 15.518  & 2.453  & 16.610  & 4.779  & 16.596  & 4.260  & 17.473  & 5.816   \\
SeqMSM&6.647  & 2.874  & 6.933  & 3.111  & 7.720  & 4.129  & 7.804  & 3.801  & 9.617  & 5.262  & 10.120  & 6.041  & 12.537  & 8.422   \\
BalanceNet &\textbf{5.356}  & 2.040  & \textbf{5.374}  & 2.078  & \textbf{5.396}  & \textbf{2.116}  & \textbf{5.422}  & \textbf{2.150}  & \textbf{5.401}  & \textbf{2.086}  & \textbf{5.378}  & \textbf{2.083}  & \textbf{5.380}  & \textbf{2.068}   \\
  \bottomrule
  \end{tabular}}
\end{table*}%

\bigskip 

\begin{figure}[h]
    \centering

    \subfloat[on JOBS]{
    \begin{minipage}[t]{0.5\linewidth}
    \centering
    \includegraphics[width=0.9\textwidth]{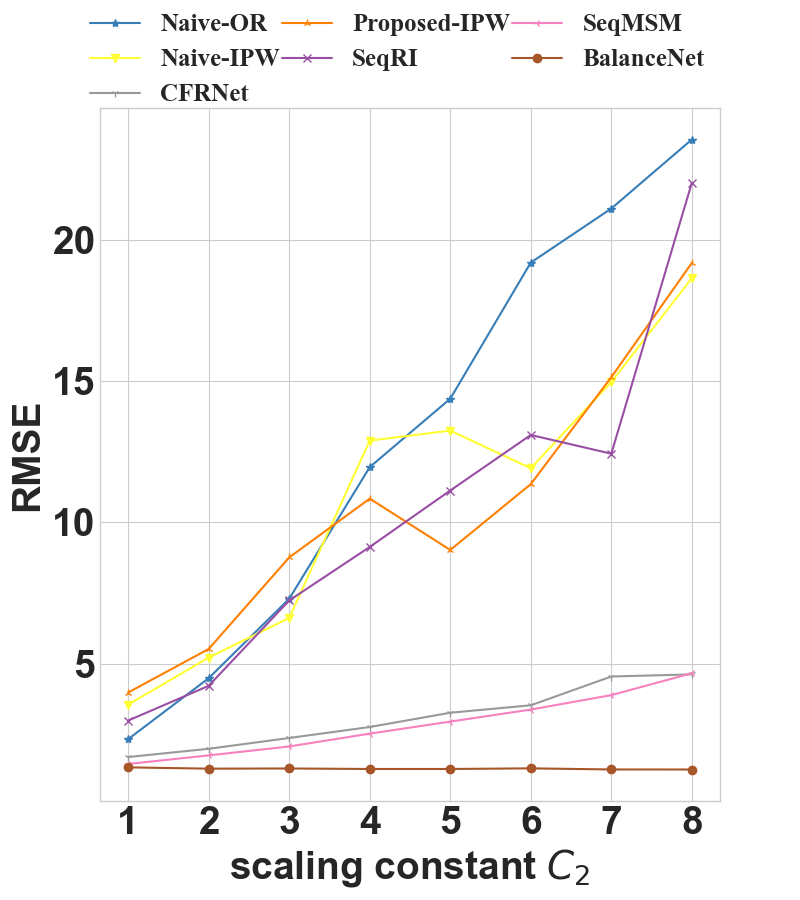}
    \end{minipage}%
    }%
    \subfloat[on IHDP]{
    \begin{minipage}[t]{0.5\linewidth}
    \centering
    \includegraphics[width=0.9\textwidth]{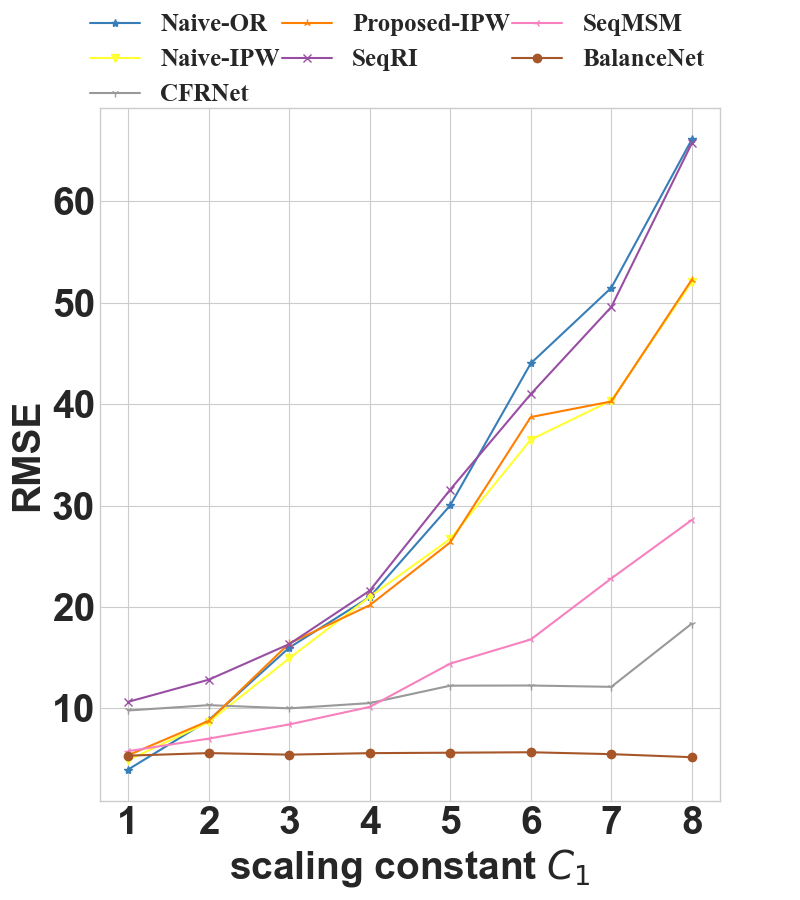}
    \end{minipage}%
    }%
    \centering
    \caption{Effects of different scaling constant on $\epsilon_{CATE}$ with missing ration $\gamma=0.15$.}
\end{figure}

\bigskip

\begin{table*}[h] 
\renewcommand \arraystretch{1.25} 
\centering
\caption{Performance comparison of variance on IHDP dataset and JOBS dataset with $\lambda_1=\lambda_2=1,C_1=3,C_2=2$.}
\resizebox{1.05 \linewidth}{!}{
\begin{tabular}{c|cc |cc|cc|cc|cc|cc|cc}
\toprule
\multicolumn{1}{c|}{JOBS} & \multicolumn{2}{c|}{$\gamma=0.05$}  & \multicolumn{2}{c|}{$\gamma=0.1$}   & \multicolumn{2}{c|}{$\gamma=0.15$}   
& \multicolumn{2}{c|}{$\gamma=0.2$}        & \multicolumn{2}{c|}{$\gamma=0.3$}    
& \multicolumn{2}{c|}{$\gamma=0.4$}
& \multicolumn{2}{c}{$\gamma=0.5$}
\\ \midrule
 Method & $\epsilon_{{CATE}_{std}}$  & $\epsilon_{{ATE}_{std}}$ & $\epsilon_{{CATE}_{std}}$  & $\epsilon_{{ATE}_{std}}$ & $\epsilon_{{CATE}_{std}}$  & $\epsilon_{{ATE}_{std}}$  & $\epsilon_{{CATE}_{std}}$  & $\epsilon_{{ATE}_{std}}$ &$\epsilon_{{CATE}_{std}}$  & $\epsilon_{{ATE}_{std}}$ &$\epsilon_{{CATE}_{std}}$  & $\epsilon_{{ATE}_{std}}$ & $\epsilon_{{CATE}_{std}}$  & $\epsilon_{{ATE}_{std}}$    
 \\ \midrule

Naive-OR &1.623  & 0.681  & 1.359  & 0.938  & 1.180  & 0.914  & 1.633  & 1.307  & 0.877  & 0.792  & 0.805  & 0.762  & 0.881  & 0.726   \\
Naive-IPW &3.573  & 1.846  & 3.191  & 1.784  & 2.155  & 1.148  & 2.251  & 1.079  & 1.680  & 0.902  & 1.552  & 0.822  & 2.487  & 1.127   \\ 
CFRNet &0.234  & 0.379  & 0.278  & 0.408  & 0.321  & 0.430  & 0.266  & 0.391  & 0.298  & 0.400  & 0.418  & 0.479  & 0.503  & 0.668   \\ 
\midrule
Proposed-IPW &5.288  & 2.641  & 3.346  & 1.741  & 2.256  & 1.471  & 2.355  & 1.119  & 1.742  & 0.986  & 2.550  & 0.972  & 1.221  & 0.925   \\ 
SeqRI &1.391  & 0.806  & 2.002  & 1.258  & 1.629  & 0.947  & 1.896  & 1.266  & 1.092  & 0.612  & 0.585  & 0.647  & 1.063  & 0.878   \\
SeqMSM &0.137  & 0.260  & 0.139  & 0.212  & 0.174  & 0.265  & 0.196  & 0.291  & 0.253  & 0.351  & 0.278  & 0.427  & 0.274  & 0.390   \\
BalanceNet &\textbf{0.054}  & \textbf{0.171}  & \textbf{0.055}  & \textbf{0.158}  & \textbf{0.055}  & \textbf{0.167}  & \textbf{0.050}  & \textbf{0.155}  & \textbf{0.077}  & \textbf{0.178}  & \textbf{0.059}  & \textbf{0.141}  & \textbf{0.076}  & \textbf{0.199}   \\

  \midrule

  \midrule
  \multicolumn{1}{c|}{IHDP}   & \multicolumn{2}{c|}{$\gamma=0.05$}  & \multicolumn{2}{c|}{$\gamma=0.1$}   & \multicolumn{2}{c|}{$\gamma=0.15$}   
& \multicolumn{2}{c|}{$\gamma=0.2$}        & \multicolumn{2}{c|}{$\gamma=0.3$}    
& \multicolumn{2}{c|}{$\gamma=0.4$}
& \multicolumn{2}{c}{$\gamma=0.5$}
\\ \midrule
 Method & $\epsilon_{{CATE}_{std}}$  & $\epsilon_{{ATE}_{std}}$ & $\epsilon_{{CATE}_{std}}$  & $\epsilon_{{ATE}_{std}}$ & $\epsilon_{{CATE}_{std}}$  & $\epsilon_{{ATE}_{std}}$  & $\epsilon_{{CATE}_{std}}$  & $\epsilon_{{ATE}_{std}}$ &$\epsilon_{{CATE}_{std}}$  & $\epsilon_{{ATE}_{std}}$ &$\epsilon_{{CATE}_{std}}$  & $\epsilon_{{ATE}_{std}}$ & $\epsilon_{{CATE}_{std}}$  & $\epsilon_{{ATE}_{std}}$    
 \\ \midrule 
 Naive-OR &1.362  & 1.254  & 1.689  & 1.532  & 2.272  & 2.252  & 3.954  & 3.430  & 5.975  & 4.222  & 6.509  & 6.020  & 5.803  & 4.631   \\
Naive-IPW &3.466  & 2.272  & 2.693  & 1.454  & 5.308  & 5.637  & 5.087  & 4.442  & 9.561  & 6.296  & 16.324  & 8.655  & 7.348  & 6.190   \\
CFRNet &3.995  & 1.102  & 3.863  & 0.966  & 5.633  & 1.704  & 5.008  & 2.148  & 5.305  & 2.843  & 9.729  & 4.651  & 6.429  & 3.278   \\
\midrule
Proposed-IPW &4.868  & 3.324  & 3.419  & 2.243  & 4.170  & 2.841  & 5.409  & 3.683  & 6.045  & 3.118  & 5.372  & 3.900  & 8.985  & 5.707   \\
SeqRI&2.665  & 1.694  & 2.715  & 1.744  & 2.862  & 1.684  & 3.159  & 2.397  & 3.851  & 3.456  & 4.263  & 3.164  & 5.400  & 4.431   \\
SeqMSM&1.607  & 1.462  & 1.580  & 1.619  & 1.998  & 2.465  & 1.693  & 2.222  & 3.438  & 3.769  & 3.536  & 3.702  & 6.350  & 6.457   \\
BalanceNet &\textbf{1.418}  & \textbf{0.144}  & \textbf{1.427}  & \textbf{0.102}  & \textbf{1.420}  & \textbf{0.103}  & \textbf{1.413}  & \textbf{0.122}  & \textbf{1.415}  & \textbf{0.121}  & \textbf{1.419}  & \textbf{0.119}  & \textbf{1.432}  & \textbf{0.115}   \\
  \bottomrule
  \end{tabular}}
\end{table*}%

\bigskip

\begin{table*}[h]
\centering
\caption{Performance comparison under different imbalance penalty $\lambda_1=\lambda_2=\lambda$ on IHDP dataset and JOBS dataset with missing ration $\gamma=0.15$.}
\resizebox{0.97 \linewidth}{!}{
\begin{tabular}{c|cc|cc|c|cc|cc}
\toprule
\multicolumn{1}{c|}{JOBS}  & \multicolumn{2}{c|}{CFRNet}   & \multicolumn{2}{c|}{BalanceNet} &\multicolumn{1}{c|}{IHDP} & \multicolumn{2}{c|}{CFRNet}   & \multicolumn{2}{c}{BalanceNet}
\\ \midrule
 Imbalance Penalty ($\lambda$) & $\epsilon_{{CATE}}$  & $\epsilon_{{ATE}}$ & $\epsilon_{{CATE}}$  & $\epsilon_{{ATE}}$ & Imbalance Penalty ($\lambda$) & $\epsilon_{{CATE}}$  & $\epsilon_{{ATE}}$ & $\epsilon_{{CATE}}$  & $\epsilon_{{ATE}}$    
 \\ \midrule
$\lambda=0.2$ &2.042  & 1.196  & \textbf{1.302}  & \textbf{0.262}  &$\lambda=0.2$ &10.143  & \textbf{1.943}  & \textbf{5.430}  & 2.158 \\
$\lambda=0.5$ &2.029  & 1.322  & \textbf{1.308}  & \textbf{0.265}   &$\lambda=0.5$ &11.515  & 2.969  & \textbf{5.417}  & \textbf{2.130}  \\
$\lambda=1.0$ & 1.991  & 1.335  & \textbf{1.307}  & \textbf{0.280}   &$\lambda=1.0$ &12.127  & 2.503  & \textbf{5.373}  & \textbf{2.101}   \\
$\lambda=1.5$&2.041  & 1.392  & \textbf{1.302}  & \textbf{0.273}    &$\lambda=1.5$&11.090  & 2.390  & \textbf{5.349}  & \textbf{2.072}  \\
$\lambda=2.0$&2.013  & 1.369  & \textbf{1.302}  & \textbf{0.283}   & $\lambda=2.0$ &10.720  & 2.652  & \textbf{5.388}  & \textbf{2.130}   \\
$\lambda=3.0$ &2.006  & 1.371  & \textbf{1.291}  & \textbf{0.245}  & $\lambda=3.0$ &11.482  & 3.249  & \textbf{5.361}  & \textbf{2.094} \\
$\lambda=4.0$ &1.992  & 1.363  & \textbf{1.296}  & \textbf{0.258}  & $\lambda=4.0$ &11.825  & 2.673  & \textbf{5.286}  & \textbf{2.024}\\
$\lambda=5.0$ &2.001  & 1.360  & \textbf{1.296}  & \textbf{0.258}  &$\lambda=5.0$ &10.750  & 2.611  & \textbf{5.299}  & \textbf{2.003}\\
  \bottomrule
  \end{tabular}}
\end{table*}%

\begin{figure}[h]
\renewcommand \arraystretch{1.25} 
    \centering
    \subfloat[on JOBS]{
    \begin{minipage}[t]{0.5\linewidth}
    \centering
    \includegraphics[width=0.9\textwidth]{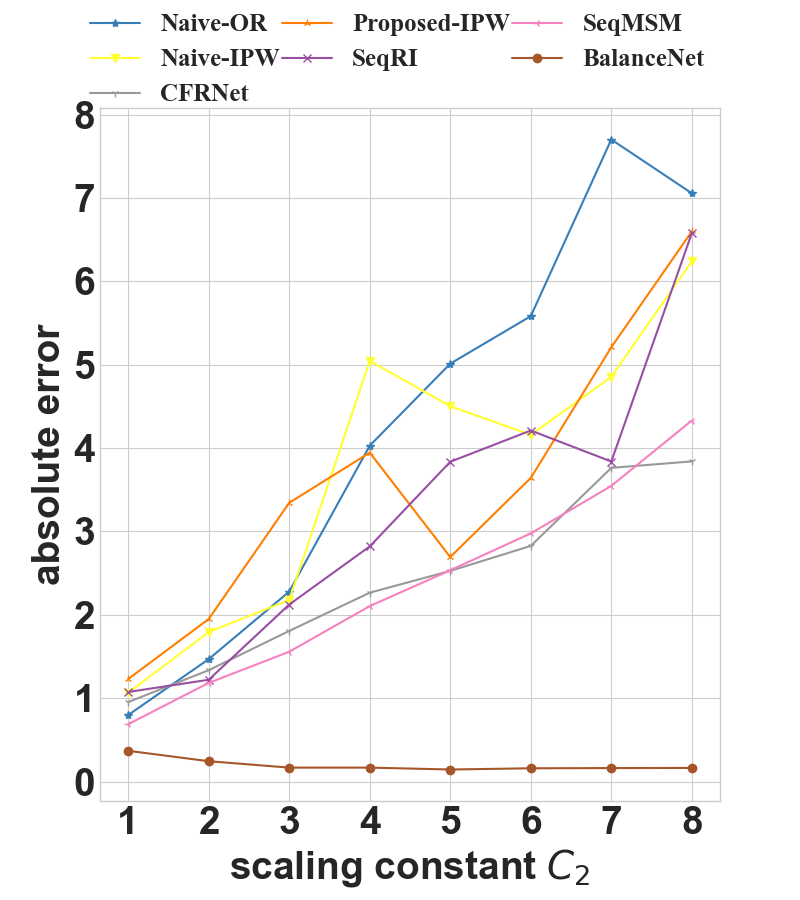}
    \end{minipage}%
    }%
    \subfloat[on IHDP]{
    \begin{minipage}[t]{0.5\linewidth}
    \centering
    \includegraphics[width=0.9\textwidth]{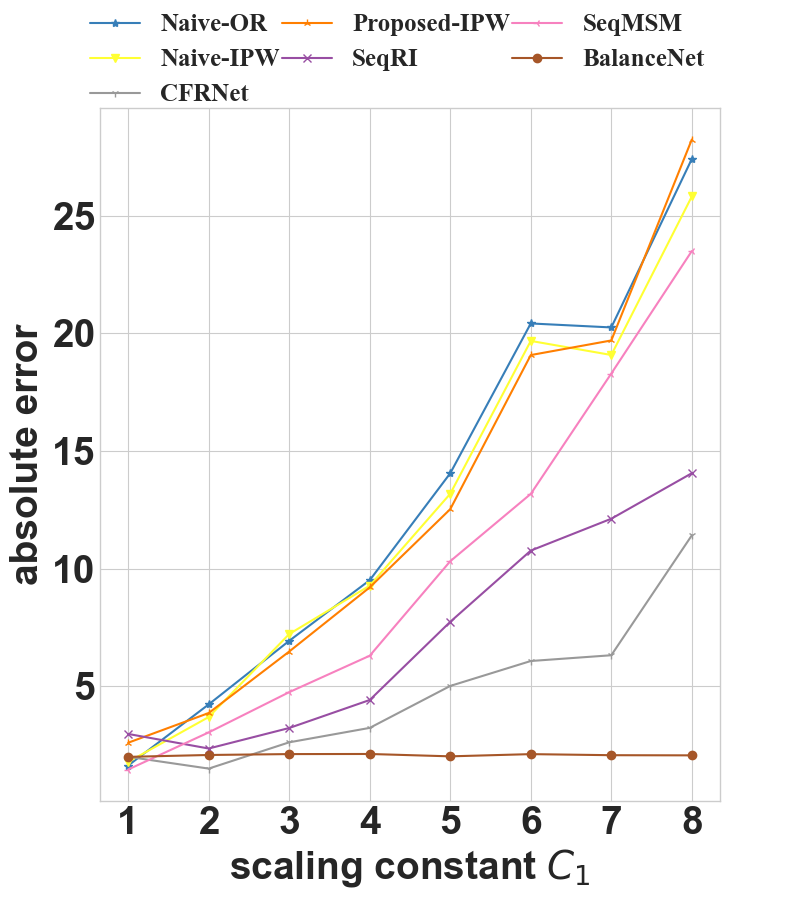}
    \end{minipage}%
    }%
    \centering
    \caption{Effects of different scaling constant on $\epsilon_{ATE}$ with missing ration $\gamma=0.15$.}
\end{figure}

\end{document}